\documentclass[12pt]{article}
\usepackage{epsfig,psfrag,amsmath,amssymb,latexsym}
\usepackage{amscd }
\usepackage{amsfonts}
\usepackage{graphicx}
\usepackage{color}
\numberwithin{equation}{section}

\begin{document}
	\title{\textbf{Is  logical analysis performed by transformers taking place in self-attention or in the fully connnected part?}}
	
\author{
	Evgeniy Shin,
	Heinrich Matzinger\thanks{
		School of Mathematics,
		Georgia Institute of Technology, Atlanta, GA 30332 \hfill\break
		Email: matzi@math.gatech.edu
	}
}

\maketitle

\begin{abstract}
Transformers architecture apply self-attention to tokens represented as vectors, before a fully connected (neuronal network) layer. These two parts can be layered many times.
Traditionally, self-attention is seen as a mechanism for aggregating information before logical operations are performed by the fully connected layer.

In this paper, we show, that quite counter-intuitively, the logical analysis can also be performed within the self-attention. For this we implement a handcrafted single-level encoder layer which performs the logical analysis within self-attention.
We then study the scenario in which a one-level transformer model undergoes self-learning using gradient descent. We investigate whether the model utilizes fully connected layers or self-attention mechanisms for logical analysis when it has the choice.
Given that gradient descent can become stuck at undesired zeros, we explicitly calculate these unwanted zeros and find ways to avoid them.
We do all this in the context of predicting grammatical category pairs of adjacent tokens in a text. We believe that our findings have broader implications for understanding the potential logical operations performed by self-attention.
\end{abstract}

\tableofcontents

\section{Introduction}
Is the logical analysis in transformers done in the self-attention part or the fully connected layers? This is the question we analyze in the current paper, specifically for the problem of predicting grammatical category pairs (or functionals there of), for adjacent tokens in the text. These category pairs contain a large part of the information from the grammatical parsing tree of each sentence. We believe that our analysis has more general meaning, then just category pairs.

Before anything else, let us present a small  example of a fully connected layer  modeling a simple logic:

\medskip
{\footnotesize
	Take $X_1,X_2,\ldots$ i.i.d Bernoulli variables with 

	$$P(X_i=1)=1-P(X_i=0)=p.$$

	Now, consider the simple logic:
	\begin{equation}
	\label{simple_logic}\left(\{X_1=1\}\cap \{X_2=1\}\right) \cup \left(\{X_3=1\}\cap \{X_4=1\}\right)\cup \left(\{X_5=1\}\cap \{X_6=1\}\right) \implies {Y=1}
	\end{equation}
	else $Y=0$. Let $\vec{X}$ be the binary vector:

	$$\vec{X}=(X_1,X_2,X_3,X_4,X_5,X_6)^t$$

	We can then represent the simple logic \ref{simple_logic}, by a fully connected neuronal network layer:

	$$Y=h(C(ReLU(B\vec{X}, {\tt bias=-1})))),$$

	where

	$$B=\left(
	\begin{array}{cc cc cc}
		1&1&0&0&0&0\\
		0&0&1&1&0&0\\
		0&0&0&0&1&1
	\end{array}
	\right)$$

	and $C=(1,1,1)$, where $h$ is the activation function with $h(x)=1$ if $x\geq 1$ and $h(x)=0$ otherwise.
}

\medskip
Now, that we have seen a first example of fully connected layer reproducing a simple logic, let us see an example which also involves self-attention.
More specially, our next example is a variation of the next word prediction task:

\medskip
{\footnotesize
The following example is given for only one reason: to show how in many cases, it is quite natural to think of a transformers architecture in the following way:
the self-attention bundles together information, which is later "logically analysed" by the fully connected layer. A little bit like in a computer with a logical printed circuit:
the circuit decides which pairs of binary numbers have to be brought together, and then applies a logical operation on them. In our case, we would think of the bringing together, as what the self-attention does, and further logical analysis would be done by fully connected layer.\\
Transformers are originally entirely trained on predicting the next word in a text. So, consider the following example:

\begin{equation}
	\label{masked_word_sentence}
	\text{Samuel receives a present. He is [MASK],}
\end{equation}
where the transformer would have to guess the masked word.  In this case, we would probably think of "Happy", or a sinonym to fill the [MASK].
Assume a simple logic with a restricted toy vocabulary, with $0,1$-encoding:\\
\begin{equation}\label{hot_encoded_words}
\begin{array}{c|c}
	\tt{word}&\tt{encoding}\\\hline\hline
	{\color{blue}\tt{receive}}&{\color{blue}(1,0,0,0,0)}\\
	{\color{green}	\tt{present}}&{\color{green}(0,1,0,0,0)}\\
	{\color{red}	\tt{pass}}  & {\color{red}(0,0,1,0,0)}\\
	{\color{yellow}	\tt{exam} }& {\color{yellow}(0,0,0,1,0)}\\
	\tt{punishment}&(0,0,0,0,1)\\
	\end{array}
	\end{equation}
We consider  the following logic:
\begin{align*}
	{\color{blue}\tt{receive}}\;{\color{green}	\tt{present}}&\implies \; happy\\
	{\color{red}	\tt{pass}}  \;	{\color{yellow}	\tt{exam} }&\implies \; happy\\
	{\color{blue}\tt{receive}}\;\tt{punishment}&\implies\; sad\\
\end{align*}
We assume that in the sentence \ref{masked_word_sentence} with the masked word, it is "clear" that it is about Samuel, so we only want to render the three logical implication above with a fully connected layer. For this we first represent each of the word pairs, $(receive,present)$, $(pass, exam)$ and $(receive punishment)$ as vector by adding
up, the words' $0,1$-hot encoding so as to get:\\
.. "{\color{blue}receive} {\color{green} present}" is encoded as:
	\begin{equation}\label{sentence1}({\color{blue}1},{\color{green}1},0,0,0)\end{equation}
	then, "{\color{red}pass} an {\color{yellow}exam}"   is encoded as:
	\begin{equation}\label{sentence2}(0,0,{\color{red}1},{\color{yellow}1},0,0)\end{equation}
Finally: "{\color{blue}receive} punishment"  encoded yields:
	\begin{equation}\label{sentence3}({\color{blue}1},0,0,0,1)\end{equation}

Let $\vec{X}$ be the vector "representing the sentence with masked word", hence one of \ref{sentence1}, \ref{sentence2} or \ref{sentence3}. We can now model "our logic"
	as a fully connected layer:
\begin{equation}\label{YCReLU}\vec{Y}=C\cdot Relu(B(\vec{X}),bias=-1)),\end{equation}
where $\vec{Y}$ is a vector with two entries: first entry represents  "happy", second entry is "sad". Always as $0$ or $1$ if the property is present. 
Also, the matrix $B$ is given as follows:
	$$
B=\left(	\begin{array}{c|cccccc}
		&receive& present&pass & exam&punishment&\\\hline
		hayppy&1         &1            &0      &0       &0     &\\
		happy&0           &0            &1      &1        &0&  \\
		sad    &1            &0           &0       &0       &1 &\\
		\end{array}\right)$$

and $C$ is given as:
	$$C=\left(
	\begin{array}{c|cccccc}
		&happy& happy&sad\\\hline
		hayppy&1         &1            &0       \\
		sad&0           &0            &1        \\
		\end{array}\right)$$

	Let $\vec{X}({\tt word})$ denote the word ${\tt word}$ represented as vector, following \ref{hot_encoded_words}. Now, for "receive a present", we
	had $\vec{X}=({\color{blue}1},{\color{green}1},0,0,0)$, and hence
	in this case:
	\begin{equation}\label{weighted_average}\vec{X}=w_1\cdot \vec{X}({\tt receive})+w_2\cdot  \vec{X}({\tt a})+w_2\cdot  \vec{X}({\tt present}),\end{equation}
	where the weights should be $w_1=1$, $w_2=0$ and $w_3=1$. Now, compare the weighted average above in \ref{weighted_average} with the formula 
	\ref{weight_av} we give for self-attention: there are the same type of weighted average. In other words, the input $\vec{X}$ we had here for our  fully connected layer
	\ref{YCReLU}, can be  simply and naturally produced by Self-attention.  Hence, we have presented 
	a natural way of first bundling the words together by self-attention, before "analyzising them logically with our fully connected layer \ref{YCReLU}". 
	The goal of this paper, is to see if this is the general approach that the transformer who self-learns choses, or if it also uses self-attention for logical operations.}

\medskip
In the current article, we do not want to predict the next token, but the grammatical category pair of adjacent tokens in the text. (So, the last example above, is different from what we do in the current article, and had the sole purpose of illustrating how natural it seems to perform the logical operation in the fully-connected layer and not in self-attention!) Finding grammatical category pairs of adjacent tokens is a sub-task of analyzing a text, the way the transformer has to do it.

We consider a sequence of  numbers denoted by $\vec{X}=(X_1,X_2,\ldots,X_M)$.  These numbers  represent the categories of words in a text, so that $X_i$ would be the category of the $i$-th token in the text. We assume that there are $N$ different categories.
We replace each digit by a double digit number obtained from the number and the number before. This leads to a vector 
$$\vec{Y}=(Y_2,Y_3,\ldots,Y_M),$$
where $Y_i=X_{i-1}X_i$ for all $i=2,\ldots,M$. We will also sometimes consider a functional of the pair $(X_{i-1},X_i)$. For this,
let $q^{\tt true}_{x_1,x_2}$ be any function from $\{1,2,\ldots ,N\}\times\{1,2,\ldots,N\}$ to $\mathbb{R}$
and put
$$Y_i=q^{\tt true}(X_{i-1},X_i),$$
for all $i=2,3,\ldots,M$.

We investigate how a one level transformer can learn to predict $\vec{Y}$ from $\vec{X}$. For this purpose, the entries of $\vec{X}$ are enriched with their position $i$. We then replace $\vec{X}$ by a matrix $X$. This means that the $i$-th column of $X$, denoted by $\vec{X}_i$, is the concatenation between the one-hot encoded $X_i$ and the one-hot encoded position $i$. The input of our one level transformer will be the matrix $X$.

We propose three hand-programmed solutions. The first and third solution uses self-attention to  add neighboring columns of $X$, then determines the  pair of categories (or functional there of) from these summed vector. The second uses the bilinear product in the self-attention to determine the $Y_j$, hence obtains the information about category pairs of adjacent tokens in $\vec{X}$ using the self-attention.

We investigate to which extend the transformer which self-learns uses one of these solution. For this we let the one-level transformer architecture self-learn with gradient-descent and analyze the outcome. One of the main problems, is that, without precaution, gradient descent gets stuck in undesirable local minimae. To remedy this, we explicitly calculate all places where gradient vanishes, for both type of solutions.
This is done in Section \ref{gradient_zero}. We are able to determine, that the only undesirable zeros are degenerate cases, which can easily be eliminated, by using Softmax and/or rescaling. With that knowledge, we than run experiment to determine, which type of solution is prefered by the self-laerning transformer. These simulation results are presented in Section \ref{simulations}.

The current article is in the tradition of articles which analyze for a given neural networks architecture, which high-dimensional function it can approximate.
Historically, early on the \cite{minsky1988perceptrons,sejnowski2018deep} showed in that many functions could not be approximated by a perceptron. This lead for a while to a freeze of AI after an initial euphoria. Later \cite{cybenko1989approximation} that multilayer perceptron is capable of distinguish data that is not linearly separable, which again raised hope.
If a certain task can be performed, by a certain architecture, but, if there is a certain ease in representing the solution to the task in the desired
architecture, i.e here one level transformer. Indeed, in appears to us, that the breakthrough like for example convolutional networks in image recognition \cite{krizhevsky2012imagenet, ciregan2012multi, dogconv}, comes about, when the architecture has a specific design which matches the structure of the problem. So, for image recognition, neural networks, where not very useful until  convolutional networks were invented \cite{ciregan2012multi}, which is basically where neural network meets traditional image recognition filters. Indeed, these convolutions are simply moving averages in little windows with go over the pixels of the image and are put one after the other.

The other aspect of our approach is that we cut the problem of text understanding by transformer into many smaller sub-parts which we then study individually each in an article. To understand a text, you first need to get which words get together in a sentence, that is for example what is the subject of a verb in a sentence. This corresponds to the grammatical parsing tree of a sentence.
We even cut this problem of parsing tree in a sentence into smaller parts, one of them consisting in recognizing category pairs for adjacent tokens. Indeed, from an information-theoretical point of view, a very large part of the information of the parsing tree of the sentence is contained in neighboring tokens' category-pairs.

As mentionned, the goal of this article is to have a one-level transformer generate $\vec{Y}$ given the matrix $X$ and study if the logical processing happens in the self-attention or the fully connected layer.
For this we will consider hand-designed one level transformers vs self-learning transformers.
We don't consider real sentences, but just generate vectors $\vec{X}$ with i.i.d digit entries.

\section{Three different hand programmed solutions}
In this section, we present our two hand programmed one-level transformers in the Subsections \ref{first_transformer} and \ref{second_transformer}.
Here, we first start by explaining, how the formula given for transformers self-attention in the original transformer paper \cite{attention_is_all_you_need}, corresponds to taking a weighted average of columns of $X$.
Typically the matrix, which encodes input to the transformer, is represented as the transpose of our matrix $X$, hence as $X^t$. In that case the $i$th row corresponds to the $i$th token encoded as a vector. 
We consider here self-attention as it is defined in the original paper \cite{krizhevsky2012imagenet}, and we are going to define it using the same fomulae.

For this, assume three matrices $q$, $k$ and $v$.
The size of $q$ and $k$ is identical, and the number of columns is equal to the dimension of the encoded tokens. (Hence, to the the number of rows of $X$, in our case, of our type of one-hot encoding for token and position the dimension is  $M+N$.)

We define
$$Q=X^t\cdot q^t$$
$$K=X^t\cdot k^t$$
\begin{equation}
\label{definition_v_V}
V=X^t \cdot v^t\end{equation}
in transformer literature $Q$ is the query, $ K$ the key and $V$ the value. Attention is then a matrix defined by
$$ATTENTION(Q,K,V)=Softmax\left(\frac{QK^t}{\sqrt{d_k}}\right)\cdot V,$$
	where $d_k$ is a scaling factor.
	In this paper, we will also use the same attention but without the Softmax. We denote it by putting a $*$
	in subscript:
	$$ATTENTION_*(Q,K,V)=\left(QK^t\right)\cdot V.$$
	Note that every row of $$ATTENTION(Q,K,V)$$ is a weighted average of lines of $X^t$. The weights are rescaled bilinear products of the lines of $X^t$. These weights are contained in the square matrix:
	$$Softmax\left(\frac{QK^t}{\sqrt{d_k}}\right).$$
	Let us see more in detail,  how this works. In our opinion to understand it easily, it is better to work with $X$ rather than its transpose.
	For that purpose, we  take the transpose of the attention. In this ways, vectors representing a token in the text are  columns of $X$.
	We find
	$$ATTENTION(Q,K,V)^t=v\cdot X\cdot  Softmax\left(\frac{KQ^t}{\sqrt{d_k}}\right),$$
	or without the Softmax:
	\begin{equation}
		\label{attention_no_softmax}
		ATTENTION_*(Q,K,V)^t=v\cdot X\cdot  KQ^t,
	\end{equation}
	where again we put an asterix $*$ in subscript to indicate that we do not use Softmax.
	Note, that when we apply the matrix 
	\begin{equation}
	\label{weight_matrix}Softmax\left(\frac{KQ^t}{\sqrt{d_k}}\right)
	\end{equation}
	to the right of $X$, this replaces $X$ by a matrix where each column is a linear combination of the columns of $X$, i.e. the i-th column of $X$, denote by $\vec{X}_i$ is replaced by a weighted average of the column before and including  the i-th:
	
	\begin{equation}\label{weight_av}w_{1i}\vec{X}_1+w_{2i}\vec{X}_2+\ldots+w_{ii}\vec{X}_i,
		\end{equation}
	where  $w_{ji}$ is the $ji$-th entry of the matrix \ref{weight_matrix} (hence found in column $i$ and row $j$).
	In other words, the weights in the weighted average \ref{weight_av} are found in the  $i$-th column of the matrix \ref{weight_matrix}.
	Now, the $j$-th entry of the $i$-th column of $K\cdot Q^t$ is obtained by taking the 
	 the $j$-th row of $K$ and calculating the dot-product  with the $i$-th column of $Q^t$. The $i$-th column of $Q^t$ equals $q\vec{X}_i$. Similarly, the $j$-th row of $K$ is $(k\vec{X}_j)^t$. Hence, we get:
	 $$w_{ji}=z\cdot exp( \vec{X}_j^t\cdot k^t q\cdot  \vec{X}_i))$$
	 where $z$ is the normalization weight:
	 $$z:=  \sum_{j=1}^i exp( \vec{X}_j^t\cdot k^t q \cdot \vec{X}_i)).$$
	 If we consider the self-Attention without Softmax, hence $ATTENTION_*(Q,K,V)$, we would have:
	  \begin{equation}\label{weights_no_softmax}w_{ji}= \vec{X}_j^t\cdot k^t q\cdot  \vec{X}_i.
	  \end{equation}
	 Note that the self-attention is only affected by the product $k^tq$ and not by the individual values of $k$ and $q$.
	 As long as their product is the same, the self-attention outcome will be the same.
	 
We are going to look at three possible solutions. The first uses self-attention to bring together columns of $X$ which are adjacent. Then, it determines the category pair using the fully connected layer. The third solution, proceeds similarly to the first, but is an improvement. The second approach is different in that it uses self-attention to determine category pairs of column pairs of $X$.  Then, it uses the skip connection followed by ReLU, to keep only category pairs from adjacent tokens.

Let us see these three possibilities.
We first start with the solution which uses self-attention to add-up adjacent columns of $X$, before analyzing the category-pairs with the fully connected layer.

\subsection{First approach: bringing the information together with self-attention}
\label{first_transformer}
Assume for example four positions. We are going to use the matrix

\begin{equation}\label{D-1}
	D_{-1}=
\left(
\begin{array}{cccc}
	0&1&0&0\\
	0&0&1&0\\
	0&0&0&1\\
	0&0&0&0
	\end{array}\right).
\end{equation}
(So, in general if $M$ is the number of categories in the text, the matrix $D_{-1}$ will be of dimension $M\times M$ and $D_{ij}=1$ if $i=j-1$ and $D_{ij}=0$ otherwise.)
If we multiply with a one-hot encoded position column vector from right and a positional  row-vector from left,
then we get $1$ if and only if the row vector encodes a position which is  one unit to the left of the position encoded in the column vector.
Otherwise the product: 
$$\overrightarrow{position}^t_i \cdot D_{-1} \overrightarrow{position}_j$$
 is zero.  We take the  matrix $k^tq$ to be the block matrix: 
$$k^tq=\left(\begin{array}{c|c}
	0&0\\\hline
	0&2D_{-1}\end{array}\right).$$
	For the last equality above to hold, we can take  $q$ 
	to be equal to:
	\begin{equation}\label{q1}q=\left(\begin{array}{c|c}	
	0&2D_{-1}
	\end{array}\right).\end{equation}
	and
$k$ to be in  in block notation:
	\begin{equation}\label{k1}k:=\left(\begin{array}{c|c}	
		0&I_M
	\end{array}\right),\end{equation}
	where $I_m$ represents the $M\times M$ identity matrix.
	Then, for $v$ we take the matrix which projects orthogonaly onto the space of categories, hence gets rid of the positional encoding:
	\begin{equation}\label{v1}v=\left(\begin{array}{c|c}	
			I_N&0\\
	0&0
	\end{array}\right)
	\end{equation}
	where $I_N$ is the $N\times N$ identity matrix and $v$ is a $(M+N)\times(M+N)$ matrix.
	In this manner $ATTENTION_*(Q,K,V)^t$ is the matrix obtained from $X$ by replacing the $i$-th column $\vec{X}_i$ by $2\cdot \vec{X}_{i-1}$
	and then deleting the positional encoding. (Here we use the self-attention without Softmax). We can then apply the Skip-connection which will add $\vec{X}_i$ to the $i$-th column of the Attention. In this manner,
	in each column of the thus obtained matrix, we have the sum of the current category of the token number $i$ plus twice the 0,1-encoded category of token
	$i-1$. To retrieve the 0,1-hot-encoded category  pair, we apply the matrix $B$ followed by Relu, where $B$ is:
	
	\begin{equation}\label{B}B=\left(\begin{array}{cccccccc}
		3&0&0&0&0&0&0&0\\
		0&3&0&0&0&0&0&0\\
		0&0&3&0&0&0&0&0\\
		0&0&0&3&0&0&0&0\\
		2&1&0&0&0&0&0&0\\
		2&0&1&0&0&0&0&0\\
		2&0&0&1&0&0&0&0\\
		1&2&0&0&0&0&0&0\\
		0&2&1&0&0&0&0&0\\
		0&2&0&1&0&0&0&0\\
		1&0&2&0&0&0&0&0\\
			0&1&2&0&0&0&0&0\\
				0&0&2&1&0&0&0&0\\
					1&0&0&2&0&0&0&0\\
						1&0&0&2&0&0&0&0\\
							1&0&0&2&0&0&0&0\\
							\end{array}\right)\;\;,\;\;
							\vec{bias}=\left(\begin{array}{c}
									-8\\
									-8\\
									-8\\
									-8\\
									-4\\
									-4\\
									-4\\
									-4\\
									-4\\
									-4\\
									-4\\
									-4\\
									-4\\
									-4\\
									-4\\
									-4\\
								\end{array}\right).
							\end{equation}
							
	The matrix $B$ has in the end, $M$ columns of $0$'s, so that positional encoding does not affect the outcome.
	This example is for the special case of $M=4$. In general, the dimension of $B$ will be $N^2\times(N+M)$)\\

	The above defined one level architecture, is able to reconstruct the category pairs, (or functionals there of)
	of adjacent tokens.
	For this let  $q^{\tt true}_{(x_1,x_2)}$ be any function from $\{1,2,\ldots ,N\}\times\{1,2,\ldots,N\}$ to $\mathbb{R}$
	and put
	$$Y_i=q^{\tt true}_{(X_{i-1},X_i)},$$
	for all $i=2,3,\ldots,M$.
	so that $\vec{Y}=(0,Y_2,Y_3,\ldots,Y_M)$. Let us define the matrix $C$ with one row of length $N^2$:
	\begin{equation}\label{C}C:=(q^{\tt true}_{(1,1)},q^{\tt true}_{(2,2)},q^{\tt true}_{(3,3)},q^{\tt true}_{(4,4)},q^{\tt true}_{(1,2)},q^{\tt true}_{(1,3)},q^{\tt true}_{(1,4)},q^{\tt true}_{(2,1)},q^{\tt true}_{(2,3)},q^{\tt true}_{(2,4)},\ldots,q^{\tt true}_{(4,3)}).
		\end{equation}
	Then, we have that $\vec{Y}$ is equal to:
	\begin{equation}\label{vecY*}\vec{Y}=C(ReLU(B\left[ATTENTION_*(Q,K,V)^t+X\right]+\vec{bias})),
		\end{equation}
	
	where $q,k,v,B$ are defined by \ref{q1}, \ref{k1},\ref{v1} and \ref{B}. 
	In other words, the one level transformer
	described by us here is able to "reconstruct" $\vec{Y}$!
When $Y_i=q^{\tt true}(X_{i-1},X_i)=X_{i-1}X_i$, then we have that $Y_i$ is just a linear function of $X_i$ and $X_{i-1}$.
Indeed in that case, we have $Y_i=10\cdot X_{i-1}+X_i$. For linear map, no Relu and fully connected layer is needed! Indeed,
you can just replace the $2$ in the formula for $q$ by $10$, so that:

\begin{equation}\label{q1new}q=\left(\begin{array}{c|c}	
		0&10D_{-1}
	\end{array}\right).\end{equation}
	Leaving out $B$ and Relu, but leaving everything else as before, whilst have the definition of $q$ given in \ref{q1new},  we find:
	$$\vec{Y}=C\cdot \left[ATTENTION_*(Q,K,V)^t+X\right]$$
	where:
	$$C:=(0,1,2,3,4,5,6,7,8,9,0,0,0,0,0,0,0,0,0,0),$$
	where the number of $0$'s at the end are for the location part of the vectors.

\medskip
{\bf Let us give a numeric example:}

{\footnotesize We assume a text of length $4$ and $4$ categories, hence $N=M=4$.
	Now let the original text be
	$$\vec{X}=(X_1,X_2,X_3,X_4)=(1,3,2,2),$$
	we want the tranformer to produce as output the vector:
	$$\vec{Y}=(0,q^{true}_{1,3},q^{true}_{3,2},q^{true}_{2,2}).$$
	We one-hot encode each entry of $\vec{X}$ and also the position to obtain the matrix $X$ defined by
	$$X
	=
	\left(
	\begin{array}{cccc}
		1&0&0&0\\
		0&0&1&1\\
		0&1&0&0\\
		0&0&0&0\\\hline
		1&0&0&0\\
		0&1&0&0\\
		0&0&1&0\\
		0&0&0&1
		\end{array}
	\right)$$
	where the line cutting through the above matrix shows the separation between the hot encoded categories and positions.
	The categories are above the line, and the positions below. Again, the $i$-th column of $X$ represents the $0,1$ hot encoded token $X_i$
	and its position. The matrix $X$ will be the input for our transformer. When we apply Self-Attention, we simply move every column of $X$, to the right, and then $v$ deletes the part below the line,
	and multiplies what is above by a factor $2$, so that:
	$$ATTENTION_*(Q,K,V)^t=
		\left(
	\begin{array}{cccc}
		0&2&0&0\\
	0&	0&0&2\\
	0&	0&2&0\\
	0&	0&0&0\\\hline
	0&	0&0&0\\
	0&	0&0&0\\
	0&	0&0&0\\
	0&	0&0&0
	\end{array}
	\right)$$
	From there the Skip-connection simply add the original matrix $X$ so that we find:
	\begin{equation}
	\label{attention+skip}ATTENTION_*(Q,K,V)^t+X=
		\left(
	\begin{array}{cccc}
		1&2&0&0\\
		0&	0&1&3\\
		0&	1&2&0\\
		0&	0&0&0\\\hline
		1&	0&0&0\\
		0&	1&0&0\\
		0&	0&1&0\\
		0&	0&0&1
	\end{array}
	\right)\end{equation}
	Now, in the last matrix above, what is below the line in the middle of the matrix is not important, because,
	when we multiply by the matrix $B$  given in \ref{B}, this part get erased. This is thanks to the $M$ columns of $B$ on the right consisting only 
	of $0$'s. Now, let us look for example at the second to last column of the matrix \ref{attention+skip} given by $(0,1,2,0,\ldots)^t$
	This vectors tells us that the category of the token $X_3$ was the third category, whilst the token $X_2$ was second category.
	Now, when multiplying that vector $(0,1,2,0,0,0,1,0)^t$ from the right by $B$, only the twelth line of $B$, when multiplied by that vector yields five,
	or other lines will yield a smaller value, which when we add the biases, will be erased by Relu. 
	Hence, we find that
	$$ReLU\left[B\cdot 
	\left(
	\begin{array}{c}
	0\\
	1\\
	2\\
	0\\
	0\\
	0\\
	1\\
	0
	\end{array}\right)+\vec{bias}\right]=
	\left(\begin{array}{c}0\\0\\0\\
		0\\0\\0\\ 
		0\\0\\0\\ 
		0\\0\\{\color{red}1}\\ 
		0\\0\\0\\0\end{array}\right)$$
		When we multiply the vector on the right side of the last equation above with $C$ defined in \ref{C}, we obtain
		the $12$ entry of $C$, that is $q^{\tt true}_{(3,2)}=q^{\tt true}_{(X_2,X_3)}$, which is equal to $Y_3$, which is what we aim at reconstructing.
		So, this will work with every column of our attention plus skip except the first:there we cna not do anything since there is no column to the left.
		So, we find:
		$$ReLU\left(B(ATTENTION_*(Q,K,V)^t+X)\right)=
		\left( 
		\begin{array}{cccc}
			0&0&0&0\\
			0&0&0&1\\
			0&0&0&0\\
			0&0&0&0\\
			
			0&0&0&0\\
			0&1&0&0\\
			0&0&0&0\\
			0&0&0&0\\
			
			0&0&0&0\\
			0&0&0&0\\
			0&0&0&0\\
			0&0&1&0\\
			
			0&0&0&0\\
			0&0&0&0\\
			0&0&0&0\\
			0&0&0&0\\
			\end{array}
			  \right)$$
			  Finally, when multiplying the last matrix above from the right by the 1row matrix $C$ given in \ref{C}, we find:
			  $$C\cdot ReLU\left(B(ATTENTION_*(Q,K,V)^t+X)\right)=(0,q^{\tt true}_{(1,3)},q^{\tt true}_{(3,2)},q^{\tt true}_{(2,2)})=
			  (0,q^{\tt true}_{(X_1,X_2)},q^{\tt true}_{(X_2,X_3)},q^{\tt true}_{(X_3,X_4)}),$$
			  which is final output and is equal to $\vec{Y}$, which we wanted to obtain.
}

	\subsection{Second approach: using self-attention for identifying category pairs}
	\label{second_transformer}
	In this current approach, the self attention first gets all the category pairs  stored in the positional encoding part
	of the space. And this not just for nearest neighbors. It then, retrieves the nearest neighbor pairs, by applying a linear map,
	followed by skip with Relu which allows to extract the diagonal. Let us see how this works:
	Again $N$ is the number of categories. For our explanation below assume $M=4$, hence four categories  to simplify notation.
	Consider the matrix $A_2$ defined by
	\begin{equation}\label{matrix1}A_2=
	\left(
	\begin{array}{cccc}
		11&12&13&14\\
		21&22&23&24\\
		31&32&33&34\\
		41&42&43&44
		\end{array}\right)
		\end{equation}
		The matrix above is for when we want to retrieve the category pairs. If we want to retrieve a functional of the category pair,
		then we need the following matrix instead:
		\begin{equation}\label{matrix2}A_2=
		\left(
		\begin{array}{cccc}
			q^{\tt true}_{(1,1)}&q^{\tt true}_{(1,2)}&q^{\tt true}_{(1,3)}&q^{\tt true}_{(1,4)}\\
			q^{\tt true}_{(2,1)}&q^{\tt true}_{(2,2)}&q^{\tt true}_{(2,3)}&q^{\tt true}_{(2,4)}\\
				q^{\tt true}_{(3,1)}&q^{\tt true}_{(3,2)}&q^{\tt true}_{(3,3)}&q^{\tt true}_{(3,4)}\\
					q^{\tt true}_{(4,1)}&q^{\tt true}_{(4,2)}&q^{\tt true}_{(4,3)}&q^{\tt true}_{(4,4)}\\
		\end{array}\right),
		\end{equation}
		where $q^{\tt true}(.,.):\{1,2,\ldots,N\}\times \{1,2,\ldots,N\}\mapsto \mathbb{R}$ is that functional. The matrix $A_2$, will basically be used to define
		$k^tq$ in the self-attention, see \ref{q} below.\\
		For now, note that when we multiply matrix \ref{matrix1} on right and left by $01$-hot encoded category, we get the two digit number
		corresponding to the category pair! Similarly, when multiply the matrix \ref{matrix2} on the right and the left by $01$-hot encoded category, we get the functional value according to $q^{\tt ture}_{(.,.)}$ of the  category pair!f Let us see an example. First take the row vector representing the first category
		$(1,0,0,0,0)$ and then a category 3 hot encode column vector:
		$(0,0,1,0)^t$ Multiplying matrix $A_2$ with these two vectors on both sides we find
		$$
		(1,0,0,0)\cdot A_2
		\cdot
		\left(
		\begin{array}{c}
			0\\
			0\\
			1\\
			0
		\end{array}\right)
	=13.$$
	Hence, the output which is the number $13$ tells us that the vector on the left is category $1$ and the vector on the right is category $3$.
	(If we would use the matrix \ref{matrix2} instead, we would find $q^{\tt true}_{(1,3)})$ instead).
	Now, we want to use the matrix $A_2$ for the bilinear product in self attention, but having it act only on the category part of the column vectors of $X$
	and not on positional encoding. In this manner, that bilinear product,
	will tell us the category pair of the vectors of which we calculate the product. 
	The position vector gets multiplied by a $0$ matrix.
	Hence, we define in block notation:
		\begin{equation}\label{q}q=\left(\begin{array}{c|c}	
		A_2&0
	\end{array}\right).\end{equation}
	And for $k$:
	$$k=\left(\begin{array}{c|c}	
		I_N&0
	\end{array}\right),$$
	where $I_N$ represents the $N\times N$ identity matrix. Again, we do not use the Softmax.
	Hence, we find that the coefficient $w_{ji}$ from self attention to be the two digit number $w_{ji}=X_jX_i$,
	which represents the categories of the $j$-th and $i$-th token in the original text.  This when we use 
	as matrix $A_2$ the version \ref{matrix1}. If on the other hand we would use \ref{matrix2}, then we would
	obtain, $w_{ji}=q^{\tt true}_{(X_j,X_i)}$. We then chose the map $v$ to   shift all entries from positional encoding down by one unit,
	and delete everything which is written in the category part of the vectors.
In other words, we define the $(N+M)\times(N+M)$ matrix $v$ by:
		$$v=\left(\begin{array}{c|c}	
			0&0\\
		0&S
	\end{array}\right),$$
	where $S$ is the linear map, mapping the hot-encoded position $j$ onto position $j+1$, and the last position onto zero.
	Hence $S=(s_{ij})_{ij}$ where $s_{ij}=1$ iff $i=j+1$ and $s_{ij}=0$ otherwise. So, $S$ is a square $M\times M$ matrix with $1$'s in a diagonal below the main diagonal.
	When we do self-attention without Softmax with $q,k,v$ as defined here, we get using the matrix \ref{matrix1} for self-attention: 
	\begin{equation}\label{attention2}ATTENTION_*(Q,K,V)^t=
	\left(\begin{array}{ccccc}
		0&0&0&0&0\\
		0&0&0&0&0\\
		0&0&0&0&0\\
		0&0&0&0&0\\
		X_1X_1&{\color{blue}X_1X_2}&X_1X_3&X_1X_4&X_1X_5\\
			0&X_2X_2&{\color{blue}X_2X_3}&X_2X_4&X_2X_5\\
				0&0&X_3X_3&{\color{blue}X_3X_4}&X_3X_5\\
					0&0&0&X_4X_4&{\color{blue}X_4X_5}\\
					\end{array}\right)
	\end{equation}
	whilst if we would use \ref{matrix2}, we would obtain:
		\begin{equation}\label{attention2b}ATTENTION_*(Q,K,V)^t=
		\left(\begin{array}{ccccc}
			0&0&0&0&0\\
			0&0&0&0&0\\
			0&0&0&0&0\\
			0&0&0&0&0\\
			q^{\tt true}_{(X_1X_1}&{\color{blue}	q^{\tt true}_{(X_1,X_2)}}&	q^{\tt true}_{(X_1,X_3)}&	q^{\tt true}_{(X_1,X_4)}&	q^{\tt true}_{(X_1,X_5)}\\
			0&	q^{\tt true}_{(X_2,X_2)}&{\color{blue}	q^{\tt true}_{(X_2,X_3)}}&	q^{\tt true}_{(X_2,X_4)}&	q^{\tt true}_{(X_2,X_5)}\\
			0&0&	q^{\tt true}_{(X_3,X_3)}&{\color{blue}	q^{\tt true}_{(X_3,X_4)}}&	q^{\tt true}_{(X_3X_5)}\\
			0&0&0&	q^{\tt true}_{(X_4,X_4)}&{\color{blue}q^{\tt true}_{(X_4,X_5)}}\\
		\end{array}\right)
	\end{equation}
	The blue diagonal of the above matrices, is what we need to retrieve. In other words, the entries in the blue diagonal give the vector $\vec{Y}$.
	We had defined Attention without Softmax in \ref{attention_no_softmax}. In that formula, the matrix $v$ is what erases all entries in the uppper half of matrix \ref{attention2b} and \ref{attention2}, which is why they have $0$'s in their upper halfs. Before applying $v$, the blue diagonal, would be one unit higher up. On the last step of calculating \ref{attention_no_softmax}, when we apply the matrix $v$, we bring the blue diagonal down by one unit. This then allows us to
	get it out late, with the help of skip and ReLU. Let us see how this works:\\
	When you are given  a vector and want to retrieve only one entry of that vector, you can proceed as follows: a) Add a large number to
	the entry you want to retrieve. b) subtract the same large number to all entries and apply ReLU. Then,  all entries
	of the vector should become $0$ except the one you wish to retrieve.. The large number just needs to be larger than all entries of the vector.\\
In order to retrieve the blue diagonal in \ref{attention2}, we add a large constant to every entry of the blue diagonal, and then apply ReLU with a negative bias of that same size. Adding a large entry to the diagonal can be done by adding skip connection times a large constant. This is so because skip adds the original vector, in this case the positional encoding which has $1$'s in the blue diagonal and $0$'s everywhere else. Let us see and example:\\
	{\footnotesize
	
	Say we have the vector $(12,32,25,11)^t$ and we would like to retrieve the third entry. We can obtain this by subtracting to each entry $100$
	and then adding the $100$ times the canonical vector $\vec{e}_3$, where $\vec{e}_3=(0,0,1,0)^t$. After that we apply ReLU:
	So, we get:
	$$
ReLU\left[100\cdot \left(	\begin{array}{c}
		0\\
		0\\
		1\\
		0
		\end{array}\right)+
		 \left(	\begin{array}{c}
			12\\
			32\\
			25\\
			11
		\end{array}\right)-
		\left(	\begin{array}{c}
			-100\\
			-100\\
			-100\\
			-100
		\end{array}\right)
		\right]=
		\left(	\begin{array}{c}
			0\\
			0\\
			25\\
			0
		\end{array}\right)
		$$
		where we see on the right of the above equation the desired result, i.e the vector where
		we only kept the third entry and all others got replaced by $0$. Here when we apply ReLU to a vector, it means that we apply it to each entry.}\\[3mm]
		Now,  in matrix \ref{attention2}, we want to retrieve from the second column the second entry. Note that  we can follow the above
		example, by adding the positional encoding of the second position. This is the second column of $X$! (Were we leave out the upper parts of the vectors which are encoding the categories.)
		So, in other words, 
		we have
\begin{multline}
\label{ReLU100}
ReLU
\left[
100\cdot X+ATTENTION_*(Q,K,V)^t,bias=-100
\right]	\\
= \left(
\begin{array}{ccccc}
	0&0&0&0&0\\
	0&0&0&0&0\\
	0&0&0&0&0\\
	0&0&0&0&0\\
	0&X_1X_2&0&0&0\\
	0&0&X_2X_3&0&0\\
	0&0&0&X_3X_4&0\\
	0&0&0&0&X_4X_5\\
\end{array}
\right)
\end{multline}
		So, we can multiply the above matrix from the left with a row of $1$'s and we will get the diagonal, that is $\vec{Y}$. Note,
		that the skip connection adds to \ref{attention2} the matrix $X$.  So, in \ref{ReLU100}, we can pull the factor $100$ out of 
		the ReLU.  The $C$ is the matrix with only one row and containing in each entry the number $100$.
		Then, we find:
		\begin{equation}\label{YCrelu}\vec{Y}=C\cdot ReLU( X+ATTENTION_*(Q,K,\frac{V}{100})^t\;,\;bias=-1).
			\end{equation}
		Here, we took the number $100$, in case of ten categories, but in general that number simply needs to be larger than all values of $q^{\tt true}(i,j)$
		for any natural numbers $i,j\leq N$.
		
		\subsection{Third approach} \label{third_transformer}We believe that most humans when asked to design a one level transformer, which would have to produce 
		$Y_i=q^{\tt true}_{X_{i-1},X_i}$ would most likely come up with the first solution we have presented in \ref{first_transformer}. 
		That solution has a practical flaw: the matrix $B$ in the fully connected layer would need too many lines: Indeed it requires $N^2$ rows,
		where $N$ is the number of categories. I in real life transformers, most matrices are square).  Now, in the solution presented here ,we are going to leave out  the matrix
		$B$ from first solution, but instead incorporate $A_2^t$ into $v$.  Let us see the details:\\
		Again we assume  for the example we give that $N=4$
		and we work with the matrix 
		\begin{equation}\label{matrix2bis}A_2=
			\left(
			\begin{array}{cccc}
				q^{\tt true}_{(1,1)}&q^{\tt true}_{(1,2)}&q^{\tt true}_{(1,3)}&q^{\tt true}_{(1,4)}\\
				q^{\tt true}_{(2,1)}&q^{\tt true}_{(2,2)}&q^{\tt true}_{(2,3)}&q^{\tt true}_{(2,4)}\\
				q^{\tt true}_{(3,1)}&q^{\tt true}_{(3,2)}&q^{\tt true}_{(3,3)}&q^{\tt true}_{(3,4)}\\
				q^{\tt true}_{(4,1)}&q^{\tt true}_{(4,2)}&q^{\tt true}_{(4,3)}&q^{\tt true}_{(4,4)}\\
			\end{array}\right),
		\end{equation}
	
	 The matrix $A_2$, was used in our second solution in Subsection \ref{second_transformer} to define
		$k^tq$ in the self-attention.\\
	Remember, we need to obtain a bilinear product of a hot encoded category vectors according to $A_2$.

	Let us give an example again: say a vector representing  the first category
		$(1,0,0,0,0)$ and then a category 3 hot encode column vector:
		$(0,0,1,0)^t$ Multiplying matrix $A_2$ with these two vectors on both sides we find
		\begin{equation}\label{bilinear_product}
		(1,0,0,0)\cdot A_2
		\cdot
		\left(
		\begin{array}{c}
			0\\
			0\\
			1\\
			0
		\end{array}\right)
		=q^{\tt true}_{(1,3)}.\end{equation}
		Let us use the following notation $X_C$ represents the $N\times M$ matrix which constitutes the upper $N$ lines of the matrix $X$,
		hence the $0,1$-hot encoded categories. Let $X_P$ represent the low part of the matrix $X$ which represents the hot-encoded poositions. Note that
		$X_p$ is a $M\times M$ identity matrix. Let us give a quick example:\\[3mm]
	 Let us give a numeric example of $X$, $X_C$ and $X_P$:

				Now let the original text be
			$$\vec{X}=(X_1,X_2,X_3,X_4)=(1,3,2,2)$$
			we $0,1$ hot encode each entry and also the position to obtain the matrix $X$ defined by
			$$X
			=
			\left(
			\begin{array}{cccc}
				1&0&0&0\\
				0&0&1&1\\
				0&1&0&0\\
				0&0&0&0\\\hline
				1&0&0&0\\
				0&1&0&0\\
				0&0&1&0\\
				0&0&0&1
			\end{array}
			\right)$$
			Then the part $X_C$ which encodes the categories would be:
			$$X_C=
				\left(
			\begin{array}{cccc}
				1&0&0&0\\
				0&0&1&1\\
				0&1&0&0\\
				0&0&0&0\\
			\end{array}
			\right)$$
			and the position encoding part:
			$$X_p=
				\left(
			\begin{array}{cccc}
				1&0&0&0\\
				0&1&0&0\\
				0&0&1&0\\
				0&0&0&1
			\end{array}
			\right)$$
		
	Now, this time we do not want a bilinear product. So, to get the bilinear product in \ref{bilinear_product}, we instead use
	a linear map followed by Relu. First posibility is to concatenate the two vectors in \ref{bilinear_product} and then have them multiply
	$(100\cdot I_M,A_2)$ (where $I_M$ is the $M\times M$ identity matrix), then apply Relu with a biase of $-100$:
	We find:
	$$q^{\tt true}_{(1,3)}=Relu\left((100\cdot I_m,A_2)
	\cdot 
	\left(
	\begin{array}{c}
		1\\0\\0\\0\\   0\\0\\1\\0
		\end{array}
	\right),bias=-100\right)$$
	The way it works, the lower part  of the vector that is $(0,0,1,0)^t$ keeps only the $3$ column of $A_2$. Then,
	when we multiply $100\cdot I_m $ by the vector $(1,0,0,0)^t$ from the right, this adds $100$ in the first row. Then,
	the ReLU with bias $-100$ deletes all other entries except the one which had receive the $+100$. Hence, in the end we get
	the first row and third entry of $A_2$ as output.\\
	Now, let $\vec{X}_{C,i}$ denote the $i$-th column of the matrix $X_C$. Hence, $\vec{X}_{C,i}$ is the category $X_i$ which is $0,1$-hot encoded.
	As the $i$-th output for our transformer $Y_i$, we want $q^{\tt true}_{(X_{i-1},X_i)}$
	which is by definition equals to:
	$$Y_i=q^{\tt true}_{(X_{i-1},X_i)}=\vec{X}_{C,i-1}^t\cdot A_2 \vec{X}_{C,i}.$$
	Since it is one dimensional it is equal to its transpose:
	\begin{equation}\label{yi}Y_i=Y^t_i=\vec{X}_{C,i}^t\cdot A_2^t\vec{X}_{C,i-1}\end{equation}
	When we take the matrix $D_{-1}$ defined at begining of Subsection \ref{first_transformer}, and apply to $X_C$ on the right,
	we get all the columns of $X_C$ shifted by one to the right, except the first, which is replaced by $0$. Consider the product:
	\begin{equation}\label{A_2}A_2 ^t X_C\cdot D_{-1}.
		\end{equation}
	Notice that the $i$th column of the last product-matrix above is equal to 
	\begin{equation}\label{vector}A_2^t\vec{X}_{C,i-1}
		\end{equation}. According to \ref{yi}, in order to obtain $Y_i$ from the vector \ref{vector}, we need to pick
		out the entry corresponding to the category $X_i$. That means if $X_i=j$, then we take the $j$-th entry of the vector \ref{vector}
		and that is equal to $Y_i$.  To extract the $j$-th entry of \ref{vector}, whilst deleting all others, we can add the $0,1$-hot encoded category $X_i$ times
		$100$ and then apply ReLU with a bias of $-100$. Hence:
		\begin{equation}Y_i=\vec{1}Relu\left(100\cdot\vec{X}_{C,i}+A_2^t\vec{X}_{C,i-1},bias=-100\right),
			\end{equation}
		where $\vec{1}$ is a vector consisting only of $1$'s of length $N$. We can now apply the above \ref{yi}, not just to the $i$-th column of \ref{A_2},
		but to every column of it. Then, instead of getting just $Y_i$, we will get the whole $\vec{Y}$, which is to say
		$$\vec{Y}=(Y_1,Y_2,\ldots,Y_M)= \vec{1}\cdot ReLU\left(100\cdot X_C+A^t_2 X_CD_{-1},bias=-100\right)$$
		The formula on the right side of the last equation above provides the same output  a one level transformer with:
		 $$k^tq=
		\left(\begin{array}{cc}
			0&0\\
			0&D_{-1}
		\end{array}\right)$$
		whilst
		$$v=
		\left(
		\begin{array}{cc}
			A^t_2&0\\
			0&0
			\end{array}\right)
$$
and given by:
$$\vec{Y}=\vec{1}\cdot ReLU\left(100\cdot X+ATTENTION_*(Q,K,V)^t, bias=-100\right),$$
where $\vec{1}$ is again a row vector consisting only of $1$'s. (Again, the formula for the Attention without softmax is given in \ref{attention_no_softmax}). We can take the factor $100$ out of ReLU to obtain:
$$\vec{Y}=100\cdot \vec{1}\cdot ReLU\left( X+ATTENTION_*(Q,K,V\cdot 0.01)^t, bias=-1\right).$$
Note that the number hundred may need to be taken even larger if the number of categories is more then ten.
It needs to be taken larger than any entry of $A_2$.

\subsubsection{Equivalence between third and second solution}

Our second and third solutions are closely related: we are going to show, that
to a large extend, they amount to exactly to the same!
Now, in the second solution presented in Subsection \ref{second_transformer},
the matrix relating to $q^{\tt true}_{(.,.)}$, that is \ref{matrix2bis},
is used for the bilinear product in the self-attention.
Here we consider self-attention without Softmax.
The matrix $k^tq$ which we use for a binlinear products between columns of $X$, 
contain the matrix $A_2$ in the second solution in \ref{second_transformer}.  These bilinear products are then used as weights to add up columns of $X$ in a weighted average defined in \ref{weight_av}.

The first solution is very different: it uses self-attention to bring 
the $(i-1)$-column in place of the $i$-th column of $X$. Then, it adds the $i$-th column again by using the skip-connection. 
The outpout of this is then analyzed by a fully connected layer, so, the bilinear functional value of the category pairs of adjacent tokens
(accroding to $A_2$)
is found out by the fully connected layer, more specifically using the matrix $B$ followed by ReLU (see \ref{B}).
The linear map $v$ out of self-attention and the subsequent $B$ of the fully connected might be combined into one matrix $B\cdot v$. So, in our third approach, presented in Subection \ref{third_transformer}, we could leave out matrix $B$, instead packing $v$, the out-matrix of self-attention with $A_2$. Also, we could have done that using a matrix $B$ and loading $A_2$ into $B$, so into the fully connected layer.
We are going to show here, that this third approach amounts to the same as the second approach, which we find a quite astonishing fact: The third approach uses
self-attention to rewrite the $i-1$-th column of $X$ in position number $i$! The second approach instead computes in self-attention all the products
according to $A_2$ of the different columns of $X_C$ and gets as output the matrix \ref{attention2b}. this matrix is totaly different from rewriting the $(i-1)$-th column of $X$ into position $i$, as is done in the third approach!

\medskip
Consider  self-attention without Softmax given by formula \ref{attention_no_softmax}, so that:
\begin{equation}\label{AQKV}
ATTENTION_*(Q,K,V)^t=v\cdot X\cdot KQ^t=v\cdot X\cdot X^t\cdot k^tq\cdot X.
\end{equation}
Again, we decompose the matrix $X$ into the upper part which is the $0,1$-hot encoding of the categories,
and the lower part which represents the positional encoding so that
$$X=\left(\begin{array}{c}X_C\\X_P\end{array}\right),$$
where $X_p$ is the $M\times M$ identity. Let us compare the two solutions:\\[3mm]

\underline{\bf Using self-attention to obtain the bilinear products:}\\
From Subsection \ref{second_transformer}, we have:
	$$v=\left(\begin{array}{c|c}	
	0&0\\
	0&S
\end{array}\right),$$
where  $S$ is a square $M\times M$ matrix with $1$'s in a diagonal below the main diagonal.
Then,
$$k^tq=\left(
\begin{array}{cc}
A_2&0\\
0&0
\end{array}\right),$$
where $A_2$ is defined according to \ref{matrix2bis}. Using formula \ref{AQKV}, we find:
$$	ATTENTION_*(Q,K,V)^t=
\left(
\begin{array}{c}
	0\\
	SX_C^tA_2X_c
\end{array}
\right)$$
Our second transformer then adds $100$ time skip to the above attention followed by a RelU with a bias of $-100$ in order to extract the diagonal.
Hence, the output is then
\begin{equation}
	\label{diagonal}\vec{Y}=diagonal(SX_C^tA_2X_c)
\end{equation}

\underline{\bf Using self-attention to move $(i-1)$-th column of $X$ to position $i$:}\\
Note that when you take the transpose of a square matrix, you obtain the same diagonal. This is to say that the output 
in\ref{diagonal} is also equal to :
$diagonal(X_C^tA_2^tX_CS^t)$. Now, note that the matrix $D_{-1}$ (see \ref{D-1}) used in the definition of the third as first transformer is the transpose of $S$!

Hence, the ouput of the second transformer given in \ref{diagonal} is equal to 
\begin{equation}\label{second}{\text{ \tt output second transformer}}=diagonal(X_C^tA_2^tX_CD_{-1}).
	\end{equation}
Now, with the third transformer, we have:
$$v=\left(
\begin{array}{cc}
A_2^t&0\\
0&0
\end{array}\right)
$$
and
$$k^tq=\left(
\begin{array}{cc}
	0&0\\
	0&D_{-1}
\end{array}\right)$$
and hence for the third solution, we find using formula \ref{AQKV}:
\begin{equation}\label{third_attention}\text{\tt Third transformer's attention:}\;\;\;\;	ATTENTION_*(Q,K,V)^t=
\left(\begin{array}{c}
	A_2^tX_CD_{-1}\\
	0
\end{array}\right)\end{equation}
At this stage the third transformer adds $100$ times skip to \ref{third_attention} followed by Relu with a bias of  $-100$.
So, the $i$-th value obtained in this way can be described as follows: take the $i$-th column of $A_2^tX_CD_{-1}$, then extract  the entry where the $i$-th column of $X_C$ has a one.
This is equal to the $i$-diagonal element of the product $X_C^tA_2^tX_CD_{-1}$!  Compare this to \ref{second},
and you see that it is identitical to the output of the second solution!  In other words, the Second solution calculates the full matrix
$X_C^tA_2^tX_CD_{-1}$ and then extract the diagonal. The third solution proceeds very similarly, but instead adds $100$ times $X_C$
to $A_2^tX_CD_{-1}$ before Relu, (instead of multiplying by $X_C^t$ from left), so that:
\begin{equation}\label{vec1}\text{\tt Ouput third transformer:}\;\;\;\vec{1}\cdot Relu[100\cdot X_C+A_2^tX_CD_{-1},bias=-100]
	\end{equation} which amounts to the same.  One more time, why is it the same? Because,
the $i$-th diagonal entry of $X_C^tA_2^tX_CD_{-1}$ is the dot product of $\vec{X}_{C,i}$ and the $i$-th column of $A_2^tX_CD_{-1}$.
(There $\vec{X}_{C,i}$ denotes the $i$th column of $X_C$). Now the vector $\vec{X}_{C,i}$ is a $0,1$ encoded category, so it contains exactly one entry equal to $1$ and all other entries are $0$. Making the dot-product with that vector, amounts to extracting the entry of the other vector in the position
of that unique $1$ of $\vec{X}_{C,i}$. But extracting that unique entry, can also be done by adding 100 times $\vec{X}_{C,i}$ followed by Relu with a bias of $-100$.
This is the the formula \ref{vec1}, which finishes to prove that both approaches lead to the same result!

\section{Zeros for gradient descent}
\label{gradient_zero}

Deep neural network get stuck in suboptimal local minimae. In the current section, we determine
the undesirable places of vanishing gradient for our one level transformers. This allows, us in our later study to avoid them. We do it for the case when 
self-attention acts only on categories, as well as for the other case, when self-attention brings together adjacent columns of $X$. It turns
out that in our two cases, the unwanted vanishing gradient place, are only degenerate points, like having the matrices $q$ and $v$ equal to zero.
Thus they can be easily be avoided by using tools  like Softmax.

\subsection{The case where self-attention acts only on the positional encoding}

We first calculate this case without Softmax:
	Assume, that the vector $\vec{Y}$ has as entry number $i$ denoted by $y_i$, the state $X_{i-1}$ hot-encoded.
	This is a little bit simpler than what we had so far where $y_i$ would contain the  hot encoded $X_i$ and $X_{i-1}$ at
	the same time. But in terms of the gradient descent getting stuck problem,
	the calculations are about the same. By taking $y_i$ to be only the state $X_{i-1}$ hot encoded, notation becomes easier.
	So, we replace in our matrix $X$, the $i$-th column by the category part of the $i-1$-th column.
	To achieve this we can take the $k^tq$ matrix to act only on the positional encoding like  subdiagonal of $1$'s.
	so more specifically in matrix-block-notation:
	$$k=(0,I)
	$$
	where as usually $I$ represents the $M\times M$ identity Then, instead of $q$ we take
	a matrix in block notation:
	$$(0,q).$$
	So that 
	$$Q^t=(0,q)\cdot X.$$
	Let $D(-1)$ denote the square matrix which is subdiagonal meaning in the diagonal below the main diagonal it has $1$'s and 
	zero's everywhere else. 
	Hence the $j,i$th entry of $D(-1)$ is $1$ if $i=j-1$ and always $0$ otherwise. Then let the map
	$v$ only keeps the category part and gets rid of the positional encoding part:
	So, the matrix $V$ defined in \ref{definition_v_V}
	is now defined by:
	$$V^t=(v,0)\cdot X.$$
	This is simply saying that $v$ acts only on the category part of $X$ and not on 
	the positional encoding. 
	If we take
	$v=id$, where $id$ represents the identity, and we take $k=(0,id)$ and $q=(0,D(-1)$
	then clearly the system just based on self-attention, without fully connected layer, 
	spits out our $y_i$. This means, that with these choices,
	we  have that
	$$ATTENTION_*(Q,K,V)^t$$
	has as $i$-th column, the 0,1-hot encoded $X_{i-1}$.
	Now, we want to train with gradient descent this attention based transformer, without it having a fully connected layer. AND yet it gets stuck. This is why we calculate here the gradient to understand why this happening and how to prevent it.
	We show this in a simple case, where self attention bilinear product (Hence $Q$ and $K$) is obtained only from 
	positional encoding, and the linear map out of self attention keeps only the category part of the vectors.

	For this, we will have that the self-attention's query matrix $Q$ be defined in block notation by:
	$$Q^t= 
	\left(
	\begin{array}{cc}
     0&q
	\end{array}\right)\cdot X,$$
	where $q$ is a $M\times M$ matrix of coefficients which
	will have to be learnt by gradient descent and
	$$K^t=\left(
	\begin{array}{cc}
		0&I	\end{array}\right)\cdot X$$
	where $I$ is the $M\times M$ identity matrix.
	Then, we take the map $v$ to only act onto category, so that
	$$V^t=
	\left(\begin{array}{cc}v&0\end{array}\right)\cdot X$$
	where $v$ is a $N\times N$ matrix of coefficients which have to be learned
	by gradient descent. Also, here we assume that $Y_i$ is just the hot encoded $X_{i-1}$ rather,
	then the couple $X_i,X_{i-1}$.  This simplifies notation in the calculations, but the same phenomena
	appears than with our main situation in this paper, where $Y_i$ encodes the category pair of adjacent 
	tokens.
	
	So, we look at the square error:
	\begin{equation}\label{leftvecy}\sum_{i=1}^M\left(\vec{Y}_i-(v,0)\cdot (q_{1i}\vec{X}_1+q_{2i}\vec{X}_2+\ldots+q_{Mi}\vec{X}_{M})\right)^2
		\end{equation}
		which will have to be summed oever the "different texts". (In transformer-theory one speaks usualy of context, to which the transformer is applied.)
		So, Instead of having only one text $\vec{X}$, we have an i.i.d. collection of $l$ texts:
		$$\vec{X}^1,\vec{X}^2,\ldots, \vec{X}^l$$
		For each of them we calculate the square prediction error given in \ref{leftvecy}, and then build the average. Since, in general we have tousands if not millions
		of such texts, (i.e. $l$ is very large), the average of the mean square error can be considered the expectation of \ref{leftvecy}. 
		So, we try to minimize the expectation of \ref{leftvecy}, hence we minimize:
		\begin{equation}\label{ESSE}E[SSE]=E\left[\sum_{i=1}^M\left(\vec{Y}_i-(v,0)\cdot (q_{1i}\vec{X}_1+q_{2i}\vec{X}_2+\ldots+q_{Mi}\vec{X}_{M})\right)^2\right],
			\end{equation}
		by gradient descent. For gradient descent to get stuck, in need to get a place where gradient is $0$. So, we are going to calculate
		for \ref{ESSE}, what the $0$'s are ( in terms of $q$ and $v$).
		 Note that for the vector $\vec{X}_j$ only the part on categories is kept
	in our formula \ref{leftvecy}. So, to simplify notation, we will consider that the vector $\vec{X}_j$ only contain the categorical information and no positional encoding. With that we get that we can replace $(v,0)$ in formula \ref{leftvecy} by just a $N\times N$ matrix  $v$ and hence get:
		\begin{equation}\label{leftvecy2}E[SSE]=E\left[\sum_{i=1}^M\left(\vec{X}_{i-1}-v\cdot (q_{1i}\vec{X}_1+q_{2i}\vec{X}_2+\ldots+q_{Mi}\vec{X}_i)\right)^2\right],
	\end{equation}
	where we also used that in the current section the $i$th column of $Y$, that is $\vec{Y}_i$ is simply the hot encoded token number $i-1$, hence
	$\vec{X}_{i-1}$.
	Let us next take the partial derivative according to $q_{ji}$:
	$$\frac{\partial E[SSE]}{\partial  q_{ji}}=E\left[\left(\vec{X}_{i-1}-v\cdot (q_{1i}\vec{X}_1+q_{2i}\vec{X}_2+\ldots+q_{Mi}\vec{X}_i)\right)\cdot \left(v\cdot\vec{X}_j\right)\right]$$
	hence
	\begin{align*}\frac{\partial E[SSE]}{\partial  q_{ji}}=&
	E\left[\vec{X}_j^t\cdot v^t \vec{X}_{i-1}-\sum_{s=1}^Mq_{si}\vec{X}_j^tv^tv\cdot \vec{X}_s\right]=\\
&=	E\left[\vec{X}_j^t\cdot v^t \vec{X}_{i-1}\right]-\sum_{s=1}^Mq_{si}E\left[\vec{X}_j^tv^tv\cdot \vec{X}_s\right]
\end{align*}
Now, when $s\neq j$ than $\vec{X}_s$ and $\vec{X}_j$ are independent so that
\begin{equation}
	\label{EXj}
	E\left[\vec{X}_j^tv^tv\cdot \vec{X}_s\right]=E\left[\vec{X}_j^t\right]v^tv\cdot E\left[ \vec{X}_s\right]
	\end{equation}
where $E\left[ \vec{X}_s\right]=E\left[ \vec{X}_j\right]$   is the vector where every entry is equal to one over the number of categories.
Hence, formula \ref{EXj} is  simply the sum of entries of the matrix $v^tv$ divided by the square of the number $N$ of categories:
\begin{equation}\label{EX2}E\left[\vec{X}_j^tv^tv\cdot \vec{X}_s\right]=\frac{\vec{1}^t v^t v \vec{1}}{N^2},
\end{equation}
where $\vec{1}$ denotes a vector with every entry being equal to $1$.
Now, when $j=s$, then the left side of equation \ref{EX2} is the trace of $v^t v$ divided by the number of categories.
This allows us to rewrite our formula for the partial derivative of SSE as:
$$\frac{\partial E[SSE]}{\partial  q_{ji}}
=	E\left[\vec{X}_j^t\cdot v^t \vec{X}_{i-1}\right]
\;-\;\frac{\vec{1}^t v^tv\cdot \vec{1}}{N^2} \sum_{s=1,s\neq j}^Mq_{si}\;-\;\frac{q_{ji}}{N} Tr(v^tv)
$$

and hence

\begin{equation}
	\label{sorry}
	\frac{\partial E[SSE]}{\partial  q_{ji}}
=	E\left[\vec{X}_j^t\cdot v^t \vec{X}_{i-1}\right]
\;-\;\frac{\vec{1}^t v^tv\cdot \vec{1}}{N^2} \sum_{s=1}^Mq_{si}\;-
\;q_{ji}\left(\frac{Tr(v^tv)}{N}-\frac{\vec{1}^t v^tv\cdot \vec{1}}{N^2} \right)
\end{equation}
Note that the middle term in the sum on the right  side of the last inequality above does not depend on $j$!
Then, if $j=i-1$, we have:
$E\left[\vec{X}_j^t\cdot v^t \vec{X}_{i-1}\right]=Tr(v^t)/10$
and hence:
\begin{equation}
	\label{sorry2}
	\frac{\partial E[SSE]}{\partial  q_{(i-1)i}}
	=	\frac{Tr(v^t)}{N}
	\;-\;\frac{\vec{1}^t v^tv\cdot \vec{1}}{N^2} \sum_{s=1}^Mq_{si}\;-
	\;q_{(i-1)i}\left(\frac{Tr(v^tv)}{N}-\frac{\vec{1}^t v^tv\cdot \vec{1}}{N^2} \right)
\end{equation}
whilst if $j\neq i-1$, we find:
\begin{equation}
	\label{sorry3}
	\frac{\partial E[SSE]}{\partial  q_{ji}}
=	\frac{\vec{1}^tv^t\vec{1}}{N^2}
\;-\;\frac{\vec{1}^t v^tv\cdot \vec{1}}{N^2} \sum_{s=1}^Mq_{si}\;-
\;q_{ji}\left(\frac{Tr(v^tv)}{N}-\frac{\vec{1}^t v^tv\cdot \vec{1}}{N^2} \right)
\end{equation}

So, the difference, when $j\neq i-1$ is:
\begin{equation}
	\label{sorry4}
	\frac{\partial E[SSE]}{\partial  q_{(i-1)i}}-\frac{\partial E[SSE]}{\partial  q_{ji}}
	={\color{blue}	\frac{Tr(v^t)}{N}}-\frac{\vec{1}^tv^t\vec{1}}{N^2}
{\color{red}	-
	\;(q_{(i-1)i}-q_{ji})\cdot \left(\frac{Tr(v^tv)}{N}-\frac{\vec{1}^t v^tv\cdot \vec{1}}{N^2} \right)}
\end{equation}
First note that our "true solution", is with $q_{(i-1)i}=1$ and $q_{ji}=0$ for $j\neq i-1$.
So, we would as the transformer is learning, we would like the term $q_{(i-1)i}$ to grow.
For this in formula \ref{sorry4}, we see that the red term has the opposite effect: 
that is when $q_{(i-1)i}>q_{ji}$ (which is what we want), we get that the red term is negative.
Hence, when $q_{(i-1)i}$ is bigger than the other terms $q_{ji}$ for $j\neq i-1$,
the red term has as effect to push $q_{(i-1)i}$ down compared to the other terms. Which is precisely not what we want.
This is why we have used the red color. Now, the blue term in formula \ref{sorry4}, can be either positive or negative at initialization.
when it is positive. it has the desired effect of lifting $q_{(i-1)i}$ above the other $q_{ji},j\neq i$. Then the blue term hence
needs to dominate the red term, for us to get the growth of $q_{(i-1)j}$ compared to the other $q_{ji}$'s, with $j\neq i$.
Below, we have evaluated the different components of our formula \ref{sorry} order of magnitude at initialization.
For this we assume at first the entries 
of $v$ are i.i.d. normal with expectation $0$ and standard deviation $\sigma$, which is the case at 
 initialization.
So, we find  {\bf initialization}
\begin{equation}\label{I}{\color{red}\frac{Tr(v^tv)}{N}=+\sigma^2O(N)}
\end{equation}
and 
\begin{equation}\label{II}{\color{red}\frac{\vec{1}^t v^tv\cdot \vec{1}}{N^2}=+\sigma^2O(1)}
	\end{equation}
then
\begin{equation}\label{III}{\color{blue}\frac{Tr(v^t)}{N}=\pm \sigma O\left(\frac{1}{\sqrt{N}}\right)}
	\end{equation}
and
finally:
\begin{equation}\label{IV}\frac{\vec{1}^tv^t\vec{1}}{N^2}=\pm \sigma O\left(\frac{1}{N}\right)
	\end{equation}
	So, if $\sigma$ where one then the term \ref{I} dominates \ref{III}, which is not what we want. 
	So, as soon as $\sigma$ is taken any border  smaller than $O(\frac{1}{\sqrt{N}})$ we are in business.
	At initialization typically for transformers, $\sigma$ is taken as 1 of over the square root of the number
	of parameters, so if this is  at initialization we are in business. However,  $\frac{Tr(v^t)}{N}$ can be positive
	or negative with equal probability.

Next we calculate the partial derivative of $E[SSE]$ according to the $kl$-th entry of $v$.  To simplify notations,
let $E[SSE_i]$ be equal to:
$$E[SSE_i]=E\left[\left(\vec{Y}_i-(v,0)\cdot (q_{1i}\vec{X}_1+q_{2i}\vec{X}_2+\ldots+q_{Mi}\vec{X}_{M})\right)^2\right],$$
so that $E[SSE]=\sum_{i=1}^ME[SSE_i]$.
We find

\begin{multline}
\frac{\partial E[SSE_i]}{\partial v_{kl}} = \\
E\left[
\left(\vec{X}_{i-1}-v\cdot (q_{1i}\vec{X}_1+q_{2i}\vec{X}_2+\ldots+q_{Mi}\vec{X}_i)\right)
\cdot
\left(E_{kl}\cdot (q_{1i}\vec{X}_1+q_{2i}\vec{X}_2+\ldots+q_{Mi}\vec{X}_i)\right)
\right]
\end{multline}
where $E_{kl}$ designates the matrix with zero's everywhere but with exactly a unique  $1$ in the $kl$-th position.
Now the expectation of the product of independents is the product of the expectations. The vectors $\vec{X}_1,\vec{X}_2,\ldots$ are i.i..d
. So, this means if we have a product involving two such different vectors we can apply expectation to each separately.
Together with linearity of expectation we find:
\begin{align}\label{henry}&E\left[\;\;\left(v\cdot (q_{1i}\vec{X}_1+q_{2i}\vec{X}_2+\ldots+q_{Mi}\vec{X}_M)\right)
\cdot \left(E_{kl}\cdot (q_{1i}\vec{X}_1+q_{2i}\vec{X}_2+\ldots+q_{Mi}\vec{X}_M\;\;)\right)\right]=\\
&=\sum_{j_1\neq j_2} q_{j_1i}q_{j_2i}E[\vec{X}_{j_1}^t]v^t\;
 E_{kl} E[\vec{X}_{j_2}]+ \sum_{j=1}^M q_{ji}^2E\left[  \vec{X}_j^tv^t E_{kl}\vec{X}_j \right]
\end{align}
Note that $E[\vec{X}_j]$  is the vector where each entry is equal to one of the number of categories, in our case $1/10$.
Hence, we get $E[\vec{X}_{j_1}^t]v^t\;E_{kl} E[\vec{X}_{j_2}]$ is the equal to the the sum of the entries of the $k$-th row of $v$ divided by $100$"
Similarly, we have $E\left[  \vec{X}_j^tv^t E_{kl}\vec{X}_j \right]$ is equal to the $kl$-th entry ov $v$ divided by the number of classes.
These two facts lead to :
\begin{align*}&E\left[\;\;\left(v\cdot (q_{1i}\vec{X}_1+q_{2i}\vec{X}_2+\ldots+q_{Mi}\vec{X}_M)\right)
	\cdot \left(E_{kl}\cdot (q_{1i}\vec{X}_1+q_{2i}\vec{X}_2+\ldots+q_{Mi}\vec{X}_M\;\;)\right)\right]=\\
	&=\sum_{j_1\neq j_2} q_{j_1i}q_{j_2i}\frac{\vec{v}_k\cdot \vec{1}}{N^2}+ \sum_{j=1}^M q_{ji}^2\frac{v_{kl}}{N}=\\
	&(q_{1i}+q_{2i}+\ldots+q_{Mi})^2\cdot \frac{\vec{v}_k\cdot \vec{1}}{N^2}+ \sum_{j=1}^M q_{ji}^2\cdot \left(\frac{v_{kl}}{N}-\frac{\vec{v}_k\cdot \vec{1}}{N^2}\right)
\end{align*}
where $\vec{v}_k$ designates the $k$-th row of the matrix $v$, written as vector.
Next, we look at 
\begin{equation}\label{little}
E[\vec{X}_{i-1}^t\cdot E_{kl} \vec{X}_j]
\end{equation}
Again if $j$ is different from $i-1$, then we get 1 over the number of states, hence: $1/N^2$. If $j=i-1$, then we get $0$ if $k\neq l$ and 
else $1/N$ when $k=l$.
Hence, we find:
$$E\left[\left(\vec{X}_{i-1}\right)
\cdot \left(E_{kl}\cdot (q_{1i}\vec{X}_1+q_{2i}\vec{X}_2+\ldots+q_{Mi}\vec{X}_M)\right)\right]=\frac{q_{1i}+\ldots+q_{Mi}}{N^2}+
q_{(i-1)i}\left(\frac{\delta_{k,l}}{N}-\frac{1}{N^2}\right),$$
where $\delta_{k,l}$ is the Kroenecker delta meaning that symbol is $1$ if $k=l$ and is $0$ otherwise.
We are now ready to put this together with the previous formula \ref{henry} and find
\begin{multline}
\label{SSEvkl}
\frac{\partial E[SSE_i]}{\partial v_{kl}} =
\frac{q_{1i}+\ldots+q_{Mi}}{N^2}
+ q_{(i-1)i}\left(\frac{\delta_{k,l}}{N} - \frac{1}{N^2}\right) \\
-(q_{1i}+q_{2i}+\ldots+q_{Mi})^2\cdot \frac{\vec{v}_k\cdot \vec{1}}{N^2}- \sum_{j=1}^M q_{ji}^2\cdot \left(\frac{v_{kl}}{N}-\frac{\vec{v}_k\cdot \vec{1}}{N^2}\right)
\end{multline}
	
	Next we are going to calculate the places where the gradient vanishes, that is where
	\begin{equation}
		\label{SSEpartv}\frac{\partial E[SSE]}{\partial  v_{kl}}=0
		\end{equation}
	for $k=1,,2,\ldots,N$
	and 

	\begin{equation}\label{SSEpartq}	\frac{\partial E[SSE]}{\partial  q_{ji}}=0
		\end{equation}
		for all $j\leq M$.
	
	We are going to sum equation \ref{SSEvkl} over the index $l$  holding $k$ fixed and then taking the average.
	Now when we average a constant we get the same constant. Hence, when we average 
	, all the terms
	which do not depend on $l$, remain unchanged. And $v_{kl}$ averaged  over $l$ hodling $k$ fixed gives by definition $\vec{v}_k\cdot \vec{1}/N$,
	i.e. the mean of the $k$-th row of the matrix $v$.
	Hence, averaging  over $l$ whilst holding $k$ fixed
	the equation
	$$0=\frac{\sum_{i=1}^M \partial E[SSE_i]}{\partial v_{kl}}$$
	where we use expression \ref{SSEvkl} for the partial derivative  $\frac{\partial E[SSE_i]}{\partial v_{kl}}$ yields:

\begin{multline}
0=
\sum_{i=1}^M\left(\frac{q_{1i}+\ldots+q_{Mi}}{N^2}\right)
+ \sum_{i=1}^M q_{(i-1)i}\left(\frac{1}{N^2}-\frac{1}{N^2}\right) \\
- \sum_{i=1}^M(q_{1i}+q_{2i}+\ldots+q_{Mi})^2\cdot \frac{\vec{v}_k\cdot \vec{1}}{N^2}
- \sum_{i=1}^M \sum_{j=1}^M q_{ji}^2\cdot \left(\frac{\vec{v}\cdot \vec{1}}{N^2}- \frac{\vec{v}_k\cdot \vec{1}}{N^2}\right)
\end{multline}

	 	Given all the values for the $q_{ij}$'s, the last equation above uniquely determines the mean of row number $k$, that is the value of 
	 	$\vec{v}_k\cdot \vec{1}/N$:
	 	\begin{equation}\label{barv_barq}\boxed{
	 			\frac{\vec{v}_k\cdot \vec{1}}{N}=
	 	\frac{\frac{\sum_{i=1}^M(q_{1i}+\ldots+q_{Mi})}{N^2}} { \sum_{i=1}^M(q_{1i}+q_{2i}+\ldots+q_{Mi})^2\cdot \frac{1}{N}}}
	 	\end{equation}
	Now, the last above equation shows that the mean of the entries of row number $k$ of the matrix $v$,
	does not depend on $k$, and hence is the same for each row. We can use this fact whilst setting the partial derivative
	$\partial E[SSE]/partial v_{kl}$ equal to $0$. Thus we sum expression \ref{SSEvkl} over index $i$ and
	equate to $0$. 
	Assume, that $l\neq k$. Then, we get the same linear equation for $v_{kl}$ which does not depend on $l$
	or $k$, since $\vec{v}_k\cdot \vec{1}$ does not depend on $k$. Hence, all values $v_{kl}$ for which $k\neq l$ are the same.
	Furthermore, if we subtract the partial derivative of $E[SSE_i]$ with respect to $v_{kl}$ from the partial derivative with respect to$v_{kk}$, with $k\neq l$,
	we get as equation:
	$$\frac{\partial E[SSE_i]}{\partial v_{kk}}-\frac{\partial E[SSE_i]}{\partial v_{kl}}=
		q_{(i-1)i}\frac{1}{N}-
		\sum_{j=1}^iMq_{ji}^2\cdot \frac{v_{kk}-v_{kl}}{N}$$

	and hence summing the above over $i$ and setting equal to $0$, yields:
		\begin{equation}\label{deltav}\boxed{\Delta v:=v_{kk}-v_{kl}=\frac{\sum_{i=1}^Mq_{i-1}}{\sum_{i=1}^M\sum_{j=1}^Mq_{ji}^2}}
			\end{equation}
	So, in other words, the matrix $v$ has everywhere the same value, except in the diagonal, where  the terms are higher than other entries of the diagonal by a quantity $\Delta v$
	given in \ref{deltav}.  Note, that hence all the diagonal entries of $v$ are equal to each other.
	
	Next we note that if all the partial derivatives of $E[SSE]$ are zero, than
	we can set expression \ref{sorry4} equal to zero and find for any $j\neq i-1$:
	$$\Delta_q:=
	q_{(i-1)i}-q_{ji}=\frac{	\frac{Tr(v^t)}{N}-\frac{\vec{1}^tv^t\vec{1}}{N^2}}{ \frac{Tr(v^tv)}{N}-\frac{\vec{1}^t v^tv\cdot \vec{1}}{N^2}}=
	\frac{1}{\Delta v},$$
	where to obtain the last equation above we used:
	\begin{equation}\label{vt}	\frac{Tr(v^t)}{N}-\frac{\vec{1}^tv^t\vec{1}}{N^2}=\Delta v\cdot (1-\frac{1}{N}),
			\end{equation}
	
		\begin{equation}\label{vt2}\frac{Tr(v^tv)}{N}-\frac{\vec{1}^t v^tv\cdot \vec{1}}{N^2}=\Delta v^2\cdot (1-\frac{1}{N}).\end{equation}
		Equation holds \ref{vt} holds, because the matrix $v$ has all entries being the same except in the diagonal where 
		entries are higher by $\Delta v$.  So, we can write the matrix $v$ as $v=\Delta v \cdot I+v_0\vec{1}\otimes\vec{1}$
		where $\vec{1}\otimes \vec{1}$ designates the square matrix with all entries equal to $1$, and $v_0$ is the constant entries size
		of $v$ outside the diagonal, whilst $I$ is the identity matrix......
		So, in other words:
		\begin{equation}\label{deltaq=1/deltav}\boxed{\Delta_q=\frac{1}{\Delta_v}}
			\end{equation}
		and since the above equation holds for all $j\neq i-1$, this means  $q_{ji}$ does not depend on $j$, when $j\neq i$!
		Note also, that according to \ref{deltaq=1/deltav}, we have that $\Delta_q$  does not depend on $i$.
		
		Next, we set expression \ref{sorry2} and \ref{sorry3} equal to zero and then add these equation summing over $j=1,2,\ldots,M$,whilst holding
		$i$ fixed,
		and then dividing by $M$. (Hence we average.) Now, the average for a constant is the constant itself,
		and hence we find:
		\begin{equation}\label{00} 0=	\frac{\vec{1}^tv^t\vec{1}}{N^2}+\frac{1}{M}\left(\frac{Tr(v^t)}{N}-	\frac{\vec{1}^tv^t\vec{1}}{N^2}\right)
		\;-\;\frac{\vec{1}^t v^tv\cdot \vec{1}}{N^2} i\cdot \bar{q}_i\;-
		\;\bar{q}_i\left(\frac{Tr(v^tv)}{N}-\frac{\vec{1}^t v^tv\cdot \vec{1}}{N^2} \right),
		\end{equation}
		
		where $\bar{q}_i$ designates the average:
		$$\bar{q}_i=\frac{\sum_{j=1}^M q_{ji}}{M}.$$
From \ref{00}, it follows that $\bar{q}_i$ does not depend on $i$, and hence $\bar{q}_i=\bar{q}$. So, equation \ref{00} yields:
\begin{equation}\label{qbar}\bar{q}=\frac{	\frac{\vec{1}^tv^t\vec{1}}{N^2}+\frac{1}{M}\left(\frac{Tr(v^t)}{N}-	\frac{\vec{1}^tv^t\vec{1}}{N^2}\right)}{\frac{\vec{1}^t v^tv\cdot \vec{1}}{N^2} M\;+
	\;\left(\frac{Tr(v^tv)}{N}-\frac{\vec{1}^t v^tv\cdot \vec{1}}{N^2} \right),}
	\end{equation}
Let $\bar{v}$ designate the mean of each row in the matrix $v$. (Recall that we have shown these row averages to be the same for different rows).
Clearly
$$\frac{\vec{1}^tv^t\vec{1}}{N^2}=\bar{v}$$
and
$$\frac{\vec{1}^t v^tv\cdot \vec{1}}{N^2} =\bar{v}^2\cdot N$$
Using the last two equations above in \ref{00}, together with \ref{vt} and \ref{vt2}, yields
\begin{equation}\label{qbar2}
\boxed{	\bar{q}=\frac{\bar{v}+\frac{1}{M}\left(\Delta v(1-\frac{1}{N})\right)}{\bar{v}^2\cdot N\cdot M\;+
		\;\left(\Delta v^2(1-\frac{1}{N})\right),}}
\end{equation}
Note that we have proven that for every column $i$ of the matrix $q$, the average $\bar{q}_i=\bar{q}$ does not depend on $i$,
and all values in column $i$ except possibly $q_{(i-1)i}$ are equal. Furthermore $\Delta q=q_{(i-1)i}-q_{ji}$, where $j\neq i$ does not depend
on $j$ or $i$. Now, since the sum of each column of $q$ is equal, we can
	rewrite equation \ref{barv_barq} as:
	\begin{equation}\label{barvN}\bar{v}\cdot N=\frac{1}{M\cdot \bar{q}}
		\end{equation}
		where again $\bar{v}$ represents the row average of matrix $v$.  (Hence, row sum in $v$ equals column sum in $q$.) 
		With this we can replace $\bar{q}$ in \ref{qbar2} by $1/(M\cdot N\bar{v})$
		and obtain:
	
		$$	\frac{1}{\bar{v}\cdot N\cdot M}=
		\frac{\bar{v}(1+\frac{1}{M}\left(\frac{\Delta v}{\bar{v}}(1-\frac{1}{N})\right)}  {\bar{v}^2\left(N\cdot M+\frac{\Delta v^2}{\bar{v}^2}\cdot( 1-\frac{1}{N})\right)}$$
		which is leads to:
		$$1=\left(N\cdot M+\frac{\Delta v^2}{\bar{v}^2}\cdot( 1-\frac{1}{N})\right)=N\cdot M\cdot (1+\frac{1}{M}\left(\frac{\Delta v}{\bar{v}}(1-\frac{1}{N})\right)
		$$
		The last equation above can be rewritten as:
		$$ 0=\frac{\Delta v^2}{\bar{v}^2}-N \frac{\Delta v}{\bar{v}}$$
		which implies that the two solutions are
		$$\frac{\Delta v}{\bar{v}}=0$$
		and the other solution:
		$$\frac{\Delta v}{\bar{v}}=N.$$
		Note that the second solution is the row sum in each row of $v$ is equal to $\Delta v$. This implies that all elements
		in $v$ are zero, except for the diagonal, where we have $\Delta v$. This is the solution we had in mind, with $v$ being the identity times
		a scalar.\\
		Next, we rewrite equation \ref{deltav}. For this we remember that all the entries in the matrix $q$, that is all $q_{ji}$ 
		are identical except the values $q_{(i-1)i}$
		which are larger by a quantity $\Delta q$ which does not depend on $i$.
		With this we find:
		\begin{equation}\label{deltav_q}\Delta v=\frac{\bar{q}+\Delta q (1-\frac{1}{i})}{M\bar{q}^2+\Delta q^2 (1-\frac{1}{M})}
		\end{equation}
		and then use \ref{deltaq=1/deltav} to replace $\Delta v$ by $1/\Delta q$ in the last equation above to obtain:
		$$\frac{1}{\Delta q}=\frac{\bar{q}+\Delta q (1-\frac{1}{M})}{M\bar{q}^2+\Delta q^2 (1-\frac{1}{M})}$$
			which leads to 
			$$\frac{1}{\Delta q}=\frac{\bar{q}+\Delta q (1-\frac{1}{M})}{M\bar{q}^2+\Delta q^2 (1-\frac{1}{M})}.$$
			The last equation above can be rewritten as:
			\begin{equation}\label{final_barq}M\cdot\left( \frac{\bar{q}}{\Delta q}\right)^2-\frac{\bar{q}}{\Delta q}=0
				\end{equation}
			and hence we have two solutions. First:
			\begin{equation}\label{0}\frac{\bar{q}}{\Delta q}=0
				\end{equation}
			and the second:
			\begin{equation}\label{1}\frac{M\cdot \bar{q}}{\Delta q}=1,
				\end{equation}
			which means that the sum $\sum_{j\leq M} q_{ji}$
			is equal to $\Delta q$. This is only possible if all the values $q_{ji}$ are zero,
			except $q_{(i-1)i}$, which is the solution we had in mind.
		
			\subsection{Solving the problem with Softmax, case when self-attention acts only on positional encoding}
			We are going to show that when using Softmax, there is no place where gradient can get stuck and there is only one solution,
			which is our solution. Now, when you use Softmax then you work under the constrain:
			\begin{equation}\label{sumq=1}\sum_{j=1}^Mq_{ji}=1
				\end{equation}
			and all the $q_{ji}\geq 1$.
			So, we just need to check that under the constrain \ref{sumq=1},
			we do not encounter a local minima under the constrain. So, we can not put
			the partial derivatives $\frac{\partial E[SSE]}{\partial q_{ji}}$ equal to zero.
			Instead, the condition is that the gradient with respect to the $q_{ji}$'s is colinear with the gradient
			of the sum $\sum_{j=1}^Mq_{ji}$. That gradient is the vector consisting only of ones.
			The gradient according to the $q_{ji}$'s, to be colinear with that vector of $1$s,
			has simply to have all entries equal. Which is the same as putting the differences equal to $0$.
			Hence, for all $j\neq i-1$, we have to set the difference given in \ref{sorry4} equal to $0$.
			This is what we did in the previous subsection and we obtained that $\Delta q=1/\Delta v$ in \ref{deltaq=1/deltav},
			which hence remained valid here. This was also possible because there is here  no constain
			on the $v_{kl}$'s, and hence the same equations remained valid for the $v_{kl}$.  Among others, the fact
			that the $v_{kl}$'s are equal everywhere except on the diagonal, where they are larger by a quantity $\Delta v$, which 
			does not depend on the column.  Equations \ref{deltav} and \ref{barv_barq}, we obtained setting these partial derivatives 
			of $E[SSE]$ with respect to the $v_{kp}$'s equal to $0$,
			and hence again  these equations remain valid in the current case. Now, equation \ref{deltav} was shown to be equivalent to
			\ref{deltav_q}, which in terms implied \ref{final_barq}, which hence remains valid here. But,  \ref{final_barq}
			was shown to imply the two possible solutions \ref{0} and \ref{1}. With $\sum_{j\neq i-1}q_{ji}=1$,
			we have that \ref{0} is impossible. Also, then \ref{1} implies $\Delta q=1$. This means that
			$q_{ji}=0$ for all $j\neq i-1$, but $q_{(i-1)i}=1$.
			So, $\Delta q=1$, and hence since \ref{deltaq=1/deltav} is still valid, we also have $\Delta v=1$.
			Now, the still valid \ref{barv_barq}, then implies $\bar{v}=1/N$ and hence the sum of each line in $v$ is equal to $1$.
			Together, with $\Delta v=1$, implies that all entries of $v$ are $0$ except in diagonal where they are $1$.
			Hence, the only possible minimum of $SSE$ under constrain $\sum_{j\leq M}q_{ji}=1$
			is:
			$$v=I$$
			where $I$ represents the $N\times N$ identity matrix,
			and $q_{ji}=0$ for all $j\leq M$ with $j\neq i-1$, whilst $q_{(i-1)i}=1$. {\bf So, with softmax, gradient descent minimizing $E[SSE]$ 
			can not get stuck in a local minimum, but finds the correct" solution!}
			
			\subsection{Gradient zero for Self-attention not having access to positional encoding}
			Here we look at extreme situation, where the $k$ and $q$ matrix only have access to the categories, hence are determining
			a functional of category pairs.
			Then, the linear map $v$ in the self attention afterwards, only accesses the positional encoding part, where the weights $w_{ij}$ 
			
			of self-attention given in \ref{weights_no_softmax} are stored.  Finally, we take the elements
			on the diagonal out . (In our solution proposed in Subsection \ref{second_transformer},
			with self-attention acting first on categories, we had then added a skip,
			and then Relu with a bias to extract that diagonal.  This can add other vanishing gradient problems,
			so here we simplify, by directly picking out the elements on the diagonal. The question is can gradient descent get stuck in zero's which do not correspond to the solution we think of? Let us calculate!
		 We assume here as before that there are $M$ positions and $N$ categories.
			We have a initial "text":
			$$\vec{X}=(X_1,X_2,\ldots,X_M)
			$$where
			$X_1,X_2,\ldots$ are i.i.d. and take values in the set $\{ 1,2,\ldots,N \}$
			which is the set of categories. We don't ask that the categories are equaly likely.
			Now, again we represent the text as a matrix $X$ where the i-th column represent
			the $i$-th token $X_i$. For this that $i$-th column of $X$ is a  concatenation of 
			the $0,1$-hot-encoded category $X_i$ and the 0,1-hot-encoded position $i$.
			We let the bilinear products obtained with $k^tq$ only act on the categories and not
			the positional  encoding. We take 
			$$Q^t=\left(\begin{array}{cc}
				q&0\\0&0
			\end{array}\right)\cdot X$$
			where $q$ is a $N\times N$ matrix to be learnt
			and
			$$K^t=\left(\begin{array}{cc}
				I&0\\0&0
			\end{array}\right)\cdot X$$
			
			where $I$ denotes the $N\times N$ identity matrix.
			We have then that $V$
			is again
			$$V^t=\left(
			\begin{array}{cc}
				0&v
			\end{array}\right)\cdot X$$
			where $v$ is a $M\times M$ matrix to be learned by gradient descent.
			We will consider:
			\begin{equation}\label{thisattention}ATTENTION_*(Q,K,V)=QK^t\cdot V,
				\end{equation}
			
			Now, we assume a given bilinear product, that is a $N\times N$ matrix, $q^{\tt true}$,
			which defines the output:
			$$Y_i=q^{\tt true}_{(X_{i-1},X_i)}$$
			In other words, the variables to be predicted $(Y_2,Y_2,\ldots, Y_M$ are  a "bilinear function of the consecutive category pairs".
		We have shown how to hand-program a transformer, which would spit out
			the $Y_{i}$s. It can be done in the format we have described in this subsection, by taking $q=q^{\tt true}$ and
			the map $v$ having ones in the  sub-diagonal of $1$'s and all other entries $0$. In Subsection \ref{second_transformer},
			where we presented that hand programmed solution, we pick the diagonal of the Attention, by applying ReLU after a skip connection. 
			This can lead to additional problems with undesired vanishing gradient, since for negative values of $x$, we have that $ReLU(x)$ has derivative $0$.
			We want to avoid this, and , instead of using ReLU, we pick the diagonal directly. This means, that our "predicted" $\hat{Y}_i$ is 
			 obtained as the $i$-th entry of the row number $i$ of \ref{thisattention}. In the present case, this leads to the simple formula:
			 $$\hat{Y}_i:=\sum_{j=1}^i  v_{ij}q_{(X_j,X_i)} .$$
		Next, we want to calculate  the 
			gradient of Sum of Square Error.
			First, if we would have only one "text":
			$$SSE_i=(Y_i-\hat{Y}_i)^2=\left(Y_i-\sum_{j=1}^N v_{ij}q_{(X_i,X_j)} \right)^2=\left(q^{\tt true}_{(X_{i-1},X_i)}-\sum_{j=1}^N  v_{ij}q_{(X_j,X_i)} \right)^2$$
			So, we will have many texts, so we will take the expectation of the above $SSE_i$. We want to calculate the partial derivative.
			Let us start with $\partial v_{si}$:
			$$2\cdot \frac{\partial  E[SSE_i] }{\partial v_{is}}=
		-E\left[	\left(q^{\tt true}_{(X_{i-1},X_i)}-\sum_{j=1}^N  v_{ij}q_{(X_j,X_i)} \right)\cdot q_{(X_s,X_i)}\right]$$
		and hence
			\begin{equation}\label{star}2\cdot \frac{\partial  E[SSE_i] }{\partial v_{is}}=
		-E\left[q^{\tt true}_{(X_{i-1},X_i)}q_{(X_s,X_i)}\right]+
		\sum_{j=1}^N v_{ij}E\left[q_{(X_j,X_i)} \cdot q_{(X_s,X_i)}\right].
		\end{equation}
		Note that $X_1,X_2,\ldots$ are i.i.d. so as long as $s,i,j$ are all different from each other,
		then the expectation below is always equal, meaning does not depend on the values of $i$, $j$ or $s$:
		$$E\left[q_{(X_j,X_i)} \cdot q_{(X_s,X_i)}\right]=E\left[q_{(X_1,X_3)} \cdot q_{(X_2,X_3)}\right].$$
		Using this and	setting the Equation \ref{star} equal to $0$ for all $s=1,2,3,\ldots,i-2$ we find:
			
			\begin{align}
			&\label{soap}E\left[q^{\tt true}_{(X_{1},X_3)}q_{(X_2,X_3)}\right]=\\
			&=E\left[q_{(X_{1},X_3)}q_{(X_2,X_3)}\right]\cdot \sum_{j\neq i}  v_{ij}+v_{is} \left(E[q^2_{(X_1,X_2)}]-
			E\left[q_{(X_{1},X_3)}q_{(X_2,X_3)}\right]\right)+v_{ii}E\left[q_{(X_{2},X_2)}q_{(X_1,X_2)}\right]
			\end{align}
			
			Next we can also find the equation for $s=i-1$ by setting \ref{star} equal to zero to find:
				\begin{align}
				&\label{soap2}E\left[q^{\tt true}_{(X_{i-1},X_i)}q_{(X_{i-1},X_i)}\right]=E\left[q^{\tt true}_{(X_{1},X_2)}q_{(X_{1},X_2)}\right]\\
				&=E\left[q_{(X_{1},X_3)}q_{(X_2,X_3)}\right]\cdot \sum_{j\neq i} v_{ij}+v_{i(i-1)} \left(E[q^2_{(X_1,X_2)}]-
				E\left[q_{(X_{1},X_3)}q_{(X_2,X_3)}\right]\right)+v_{ii}E\left[q_{(X_{2},X_2)}q_{(X_1,X_2)}\right]
			\end{align}
			Finally for $s=i$, we find the equation:
			\begin{align}\label{soap3}
				&E\left[q^{\tt true}_{(X_{i-1},X_i)}q_{(X_{i},X_i)}\right]=E\left[q^{\tt true}_{(X_{1},X_2)}q_{(X_{2},X_2)}\right]\\
				&=E\left[q_{(X_{1},X_2)}q_{(X_2,X_2)}\right]\cdot \sum_{j}  v_{ij}+v_{ii}\left(E[q^2_{(X_1,X_1)}]-
				E\left[q_{(X_{1},X_2)}q_{(X_2,X_2)}\right]\right)
			\end{align}
			Note that equation \ref{soap}, tells us that all values $v_{is}$ for $s\neq i, i-1$ are equal, when
			$$\Delta Q= E[q^2_{(X_1,X_2)}]-
			E\left[q_{(X_{1},X_3)}q_{(X_2,X_3)}\right]\neq 0,$$
			which we will see is going to be the case, except for degenerate $q^{\tt true}$.
			We have that \ref{soap}, \ref{soap2} and \ref{soap3} constitute a linear system of equation for the three quantities
			$v_{ii}$, $v_{i(i-1)}$ and $\sum_{j\neq i, i-1}v_{ij}$. The system determines these values uniquely. Hence they do not depend on $i$,
			since the coefficient of the system of linear equation do not! This means that 
			$$v_{11}=v_{22}=v_{33}=\ldots=v_{ii}=\ldots=v_{NN},$$
			and
			$$v_{21}=v_{32}=\ldots=v_{i(i-1)}=\ldots=v_{N(N-1)}$$
			furthermore $\sum_{j\neq i,i-1} v_{ij}$ does not depend on $i$!

Next we want to calculate the partial derivative of expected $SSE_i$ with respect to $q_{r,t}$:
			we find:
			$$\frac{\partial  E[SSE_i] }{\partial q_{rt}}=
			-E\left[	\left(q^{\tt true}_{(X_{i-1},X_i)}-\sum_{j=1}^i  v_{ij}q_{(X_j,X_i)} \right)\cdot \sum_{k=1}^i  v_{ik}\delta_{(X_k=r,X_i=t)} \right]$$
			so setting the last equation equal to 0 yields:
			\begin{equation}\label{v}\sum_{k} v_{ik}E\left[	q^{\tt true}_{(X_{i-1},X_i)}\delta_{(X_k=r,X_i=t)}\right]=
		\sum_{j}  \sum_{k}  v_{ij}v_{ki}E\left[q_{(X_j,X_i)}\delta_{(X_k=r,X_i=t)}\right]
			\end{equation}
			Note when $k=i-1$,
			then
			\begin{equation}\label{uno}E\left[	q^{\tt true}_{(X_{i-1},X_i)}\delta_{(X_k=r,X_i=t)}\right]=P(X_1=r)\cdot P(X_1=t)q^{\tt true}_{(r,t)}
				\end{equation}
			and when $kneqi,i-1$,then
			\begin{equation}\label{duo}E\left[	q^{\tt true}_{(X_{i-1},X_i)}\delta_{(X_k=r,X_i=t)}\right]=P(X_1=r)\cdot P(X_1=t)E[q^{\tt true}_{(X_1,t)}]
				\end{equation}
				and when $k=i$, then $X_k=r$ and $X_i=t$ is impossible unless $r=t$,
				so that for $k=i$, we have:
					\begin{equation}\label{tres}E\left[	q^{\tt true}_{(X_{i-1},X_i)}\delta_{(X_i=r,X_i=t)}\right]= P(X_1=t)\cdot E[q^{\tt true}_{(X_1,t)}]\cdot \delta_{r=t}
				\end{equation}
			When $k=j\neq i$,
			then
			\begin{equation}\label{quatro}E\left[q_{(X_j,X_i)}\delta_{(X_k=r,X_i=t)}\right]=P(X_1=t)\cdot P(X_1=r)\cdot q_{(r,t)}
				\end{equation}
				When $k=j=i$,
				then
				\begin{equation}\label{cinqo}E\left[q_{(X_j,X_i)}\delta_{(X_k=r,X_i=t)}\right]=\delta_{r=t} P(X_1=t)\cdot q_{(t,t)}
				\end{equation}
				
			and when $k\neq j$ and $k,j\neq i$,
			then
			\begin{equation}\label{seis}E\left[q_{(X_j,X_i)}\delta_{(X_k=r,X_i=t)}\right]=P(X_1=r)\cdot P(X_1=t)\cdot E[q_{(X_1,t)}]
				\end{equation}
				then for $k\neq j=i$
				we find
					\begin{equation}\label{siete}E\left[q_{(X_j,X_i)}\delta_{(X_k=r,X_i=t)}\right]=P(X_1=r)\cdot P(X_1=t)\cdot q_{(t,t)}
				\end{equation}
				Also the last case is when $j\neq k=i$
				and we find in that case:
				\begin{equation}\label{ocho}E\left[q_{(X_j,X_i)}\delta_{(X_k=r,X_i=t)}\right]=\delta_{r=t}\cdot P(X_1=t)\cdot E[q_{(X_1,t)}]
				\end{equation}
				Now, when we apply \ref{uno}, \ref{duo}, \ref{tres}, \ref{quatro}, \ref{cinqo},\ref{seis} to  \ref{v}, we find:
				
\begin{multline}
\label{big}
v_{i(i-1)}\cdot q_{(r,t)}^{\tt true}
+ E[q^{\tt true}_{(X_1,t)}]\left(\sum_{k\neq i,i-1} v_{ik}+\frac{\delta_{r=t}v_{ii}}{P(X_1=t)}\right) = \\
= E[q_{(X_1,t)}]\sum_{j\neq i} \sum_{k\neq i} v_{ij}v_{ik}+v_{ii}\sum_{k\neq i} v_{ik} q_{(t,t)}
+ v_{ii}\sum_{j\neq i} v_{ij}\delta_{r=t}\frac{E[q(X_1,t)]}{P(X_1=t)} \\
+ (q_{(r,t)}-E[q_{(X_1,t)}])\sum_{j\neq i} v_{ij}^2+\frac{q_{(t,t)}\delta_{r=t}v_{ii}^2}{P(X_1=t)}
\end{multline}

Now, we are going to make a weighted average of the equation above, summing according to the index $r$ and multiplying by $P(X_1=r)$
(the weights of the weighted average).
			This way $q^{\tt true}_{(r,t)} $becomes $E[q^{\tt true}_{(X,t)}]$ and $q_{(r,t)} $becomes $E[q_{(X,t)}]$.
			So, doing the weighted average of \ref{big} as discussed, we find:
			$$E[q^{\tt true}_{(X,t)}]\sum_{k} v_{ik}=
			E[q(X,t)]\sum_{j\neq i}\sum_{k}v_{ij} v_{ik}+v_{ii}q_{(t,t)}\sum_{k}v_{ik}$$
			hence
			\begin{equation}\label{100}\boxed{E[q^{\tt true}_{(X,t)}]=E[q(X,t)]\cdot \sum_{j\neq i} v_{ij}+v_{ii}q_{(t,t)}}
			\end{equation}
			
			Next note that for any $r_1\neq r_2$ so that $r_1,r_2\neq t$, we can subtract equation \ref{big} for $r_2$ from the same equation  but with $r_1$
			so that all the terms which are not indexed in $r$ disappear. We find:
			$$v_{i(i-1)} (q^{\tt ture}_{(r_1,t)})-q^{\tt true}_{(r_2,t)})=(q_{(r_1,t)}-q_{(r_2,t)})\sum_{j\neq i} v_{ij}^2$$
			which implies that
			\begin{equation}\label{qa}q_{(r,t)}=a(t)+\frac{v_{i,(i-1)}}{\sum_{j\neq i}v_{ij}^2}q^{\tt true}_{(r,t)}+\Delta q(t)\delta_{r=t}
				\end{equation}
			where $a=a(t)$ and $\Delta q=\Delta q(t)$ are unknown functions of $t$.
			Now, we are going back to subtracting equation \ref{big} with $r=t$ from the same equation but with a value for $r$ which is not equal to $t$.
			All terms, which are not affected by $r$ dissappear, and we get:
			$$E[q^{\tt true}_{(X,t)}]\frac{v_{ii}}{P(X_1=t)}=v_{ii}\sum_{j\neq i}v_{ij}\frac{E[q(X,t)]}{P(X_1=t)}+\Delta q \sum_{j\neq i}v^2_{ij}+q_{(t,t)}\frac{v_{ii}^2}{P(X_1=1)}$$
			Putting the last equation above together with equation \ref{100},
			yields:
			\begin{equation}\label{Deltaq}\boxed{\Delta q \sum_{j\neq i}v_{ij}^2=0}
				\end{equation}
which implies that $\Delta q=0$ unless the vector $(v_{i1},v_{i2},\ldots,v_{ii})$ has only zero's up to last entry!			
Note that from equation \ref{soap} we can see that all values for $v_{ij}$ for $j\neq i,i-1$ are equal.
So, we are going to write the vector $(v_{ij})$ as follows:
$$(v_{i1},v_{i2},\ldots,v_{ii})=(v_{i1},v_{i1},v_{i1},\ldots,v_{i1},,v_{i1}+\Delta v,v_{ii},v_{i(i+1)=v_{i1},\ldots,v_{iN}=v_{i1}})$$
We can next subtract any equation \ref{soap} from equation \ref{soap2} to obtain
$$\Delta v=
\frac{E\left[q^{\tt true}_{(X_{1},X_2)}q_{(X_{1},X_2)}\right]-E\left[q^{\tt true}_{(X_{1},X_3)}q_{(X_2,X_3)}\right]}
{ \left(E[q^2_{(X_1,X_2)}]-
	E\left[q_{(X_{1},X_3)}q_{(X_2,X_3)}\right]\right)}$$
The last equation above, together with  equation \ref{qa}, whilst assuming $\Delta q=0$, yields:
\begin{equation}\label{boxeddeltav}\boxed{\Delta v=\frac{\sum_{j\neq i}v^2_{i,j}}{v_{i,(i-1)}}}.
	\end{equation}
We apply to the last equation above the fact that all $v_{ij}$ for $j\neq i,i-1$ are equal,
and $v_{i(i-1)}=v_{i1}+\Delta v$ to get the equation:
\begin{equation}\label{vi1}v_{i1}\cdot \left(v_{i1}(N-1)+\Delta v\right)=0
	\end{equation}
It follows that either $v_{i1}=0$, which corresponds to our solution, or
\begin{equation}\label{moscito}\Delta v=-v_{i1}(N-1)
	\end{equation}
which we are going to prove is an invalid solution.\\
Note when \ref{moscito} holds,
then 
\begin{equation}\label{zero}\sum_{j\neq i}v_{ij}=0
	\end{equation}
and hence, when applying this to \ref{soap3}, we find:
\begin{equation}\label{vii}\boxed{v_{ii}=\frac{E[q^{\tt true}_{(X_1,X_2)} q_{(X_2,X_2)}]}{E[q^2_{(X_1,X_1)}]}}
\end{equation}
Now, we can use \ref{zero} with \ref{100} to find:
\begin{equation}\label{viiqtt}E[q^{\tt true}_{(X,t)}]=
v_{ii}q_{(t,t)}
\end{equation}

\subsubsection{Proving that $\sum_{j\neq i}v_{ij}=0$ is invalid}
We are going to use \ref{zero} and \ref{viiqtt} to show, that we would have an invalid solution.
We are going to use equation \ref{soap2} to show that there can be no solution with \ref{zero} except the trival where all the $v_{ij}$'s and all the $q_{(r,t)}$ are all $0$. First note that with the help of \ref{viiqtt}, we find:
\begin{equation}\label{vii}v_{ii}E\left[q_{(X_{2},X_2)}q_{(X_1,X_2)}\right]=v_{ii}E\left[v_{ii}q_{(X_{2},X_2)}q_{(X_1,X_2)}\right]=
	E\left[\; E[q^{\tt true}_{(X_1,X_2)}|X_2]\cdot q_{(X_1,X_2)}\right].
	\end{equation}
	We can always assume $v_{i(i-1)}=1$ without restriction. Indeed, if we multiply $q(.,.)$ by a constant and divide the function $v_{.,.}$ by the same
	constant the output of our transformer will be the same! Now, with the help of  equation \ref{vii} and \ref{zero}, equation \ref{soap2} becomes:
		\begin{align}
		&\label{soap2bis}E\left[q^{\tt true}_{(X_{1},X_2)}q_{(X_{1},X_2)}\right]=\\
		&= \left(E[q^2_{(X_1,X_2)}]-
		E\left[q_{(X_{1},X_3)}q_{(X_2,X_3)}\right]\right)+E\left[\; E[q^{\tt true}_{(X_1,X_2)}|X_2]\cdot q_{(X_1,X_2)}\right].
	\end{align}
	Now, we assumed that $v_{i(i-1}=1$ and we have that all $v_{ij}$ for which $j\neq i,i-1$ are equal. Then from \ref{zero}, it follows that $v_{ij}=-\frac{1}{N-2}$
	for all $j\neq i,i-1$ and hence
	$$\frac{v_{i(i-1)}}{\sum_{j\neq i}v_{ij}^2}=1=1-\frac{1}{N-1}$$
	so that \ref{qa} can be rewritten
	\begin{equation}\label{qa2}q_{(r,t)}=a(t)+\left(1-\frac{1}{N-1}\right)q^{\tt true}_{(r,t)}
		\end{equation}
	where we also used that $\Delta q(t)=0$ according to \ref{Deltaq}. We are next going to use \ref{qa2} in \ref{soap2bis}.
	Now, when we apply \ref{qa2} to $E[q^2_{(X_1,X_2)}]-
	E[q_{(X_{1},X_3)}q_{(X_2,X_3)}]$ all terms containing $a(.)$ cancel out, and we find:
	\begin{equation}\label{var}
		\left(E[q^2_{(X_1,X_2)}]-
		E\left[q_{(X_{1},X_3)}q_{(X_2,X_3)}\right]\right)=
	\left(1-\frac{1}{n-1}\right)^2\cdot	\left(E[(q^{\tt true}_{(X_1,X_2)})^2]-
		E\left[q^{\tt true}_{(X_{1},X_3)}q^{\tt true}_{(X_2,X_3)}\right]\right)
	\end{equation}
	Now, note that
	$$E\left[\; E[q^{\tt true}_{(X_1,X_2)}|X_2]\cdot a(X_2)\right]=
		E\left[\; E[q^{\tt true}_{(X_1,X_2)}\cdot a(X_2)|X_2]\;\right]=
	E[q^{\tt true}_{(X_1,X_2)}\cdot a(X_2)].$$
	This last inequality shows, that when in \ref{soap2bis}, we replace on the left side $q(X_1,X_2)$ by a combination of $q^{\tt true}(X_1,X_2)$
	and $a(X_2)$ and we do the same on the right side in the part
	$$E\left[\; E[q^{\tt true}_{(X_1,X_2)}|X_2]\cdot q_{(X_1,X_2)}\right].$$
	then again the terms containing $a(X_2)$ cancel. This  together with \ref{var} allows us to rewrite \ref{soap2bis} 
	as:
		\begin{align}
		&\label{soap2bisbis}
		E\left[q^{\tt true}_{(X_{1},X_2)}q^{\tt true}_{(X_{1},X_2)}\right]=\\
		&= \left(1=\frac{1}{n-1}\right)\cdot \left(E[(q^{\tt true}_{(X_1,X_2)})^2]-
		E\left[q^{\tt true}_{(X_{1},X_3)}q^{\tt true}_{(X_2,X_3)}\right]\right)+E\left[\; E[q^{\tt true}_{(X_1,X_2)}|X_2]\cdot q^{\tt true}_{(X_1,X_2)}\right].
	\end{align}
	Next note, that
	\begin{equation}\label{verylast}E\left[q^{\tt true}_{(X_{1},X_3)}q^{\tt true}_{(X_2,X_3)}\right]=E\left[\; E[q^{\tt true}_{(X_1,X_2)}|X_2]\cdot q^{\tt true}_{(X_1,X_2)}\right].\end{equation}
	This is so because for independents the expectation of the product is the product of the expectations. And in left side of \ref{verylast}, when we condition
	on $X_3$, the terms in the product become independent, hence:

\begin{align*}
E\left[q^{\tt true}_{(X_{1},X_3)}q^{\tt true}_{(X_2,X_3)}\right] &= \\
&= E\left[E\left[q^{\tt true}_{(X_{1},X_3)}q^{\tt true}_{(X_2,X_3)}\right]|X_3]\;\right] \\
&= E\left[E[q^{\tt true}_{(X_{1},X_3)}|X_3]\cdot E[q^{\tt true}_{(X_2,X_3)}|X_3]\;\right] \\
&= E\left[E[q^{\tt true}_{(X_{1},X_3)}|X_3]^2\;\right]
\end{align*}
but similarly
\begin{align*}
E\left[\; E[q^{\tt true}_{(X_1,X_2)}|X_2]\cdot q^{\tt true}_{(X_1,X_2)}\right] &= \\
&= E\left[\;E\left[\; E[q^{\tt true}_{(X_1,X_2)}|X_2]\cdot q^{\tt true}_{(X_1,X_2)}|X_2\right]\;\right] \\
&= E\left[ E[q^{\tt true}_{(X_1,X_2)}|X_2]\cdot E[q^{\tt true}_{(X_1,X_2)}|X_2]\;\right] \\
&= E\left[ E[q^{\tt true}_{(X_1,X_2)}|X_2]^2\;\right]
\end{align*}
which finishes to prove \ref{verylast}. Now with the help of \ref{verylast}, we obtain that \ref{soap2bisbis}, can be written as:
$$0=E[(q^{\tt true}_{(X_1,X_3)})^2]-
E\left[q^{\tt true}_{(X_{1},X_3)}q^{\tt true}_{(X_2,X_3)}\right]$$
Conditioning on $X_3$, we find that the right side of the last equation above is:
\begin{align*}
E\left[\;E[\;(q^{\tt true}_{(X_1,X_3)})^2-
q^{\tt true}_{(X_{1},X_3)}q^{\tt true}_{(X_2,X_3)}|X_3]\right] &=\\
&=E\left[\;E[\;(q^{\tt true}_{(X_1,X_3)})^2-
E[q^{\tt true}_{(X_{1},X_3)}|X_3]\cdot E[q^{\tt true}_{(X_2,X_3)}|X_3]\right] \\ &= \sum_t VAR[q^{\tt true}(X_1,t)]\cdot P(X_3=t)
\end{align*}
		The fact that the last expression on the right side of the last equation above must be zero, implies that all the variances
		$VAR[q^{\tt true}(X_1,t)]$, where $t$ is a non-random number are $0$ for every $t$. This implies that the function $q^{\tt true}_{(r,t)}$
		only depends on $t$ but not on $r$. This is not the case in any aplication we care considering. Hence, we can reject this! Hence,
		we have just proven that we can not have 
		$$\sum_{j\neq i} v_{ij}=0$$
		so that by \ref{vi1} we have instead:
		$$v_{ij}=0$$
		for all $j\neq i,i-1$. Now, we don't need to have $v_{ii}=0$. Let us see why. From equation \ref{qa},
		in the current case, with $v_{i(i-1)}=1$ and $v_{ij}=0$ for all $j\neq i,i-1$, we obtain
		$$q^{\tt ture}_{(r,t)}=a(t)+q_{(r,t)}.$$
		We can apply the last equation above to \ref{100}, to find:
		\begin{equation}\label{at}0=a(t)+v_{ii} q_{(t,t)}
			\end{equation}
			or equivalently
			$$a(t)=-\frac{v_{ii}}{1+v_{ii}}\cdot  q^{\tt true}_{(t,t)}.$$
			Now, the canonical solution is $v_{i(i-1)}=1$ all other $v_{ij}$'s are $0$ and $q_{(r,t)}=q^{true}_{(r,t)}$. Of course, you could always, in that canonical
			solution multiply the function $q(.,.)$ by a constant and divide $v(.,.)$ by the same constant, and obtain the same output. But, there is also another solution,
			where $v_{ii}$ is not $0$! For that case, again $v_{i(i-1)}=1$, all $v_{ij}=0$ for $j\neq i,i-1$, but $q_{(r,t)})=q^{\tt true}_{(r,t)}+a(t)$. Then, 
			$v_{ii}$ and $a(t)$ have to satisfy \ref{at}, and the output of the transformer will be the same! The term added when calculating 
			$\hat{Y}_i$ is first $v_{i(i-1)} a(X_i)$
will be compensated by $v_{ii} q_{(X_i,X_i)}$ according to \ref{at}.			
		
\section{Learning q,k,v,C,B}
\label{simulations}
\label{self-attention_vs_fully_connected_learning}
The goal of the current article, is to find out, if the self-trained transformer uses self-attention for some of the logical analysis,
or proceeds more like our first or third solutions. In the current section, we will train a one level transformer, and study, which of the parts is used for the "logical analysis".

We have presented so far three solutions for building a hand-programmed one level transformer to  output $Y_i=q^{\tt true}(X_{i-1},X_i)$,
for a given function $q^{\tt true}(.,.)$, when given $X$ as input. 
The first solution was presented in Subsection \ref{first_transformer}, the second solution in \ref{second_transformer} and the third in \ref{third_transformer}.
		
The first solution, presented in Subsection \ref{first_transformer} adds pairs of adjacent columns of $X$ with the help of self-attention and skip connection. Then, it  extracts the information about 
the category pair $(X_{i-1},X_i)$, with the fully connected layer, that is with matrix $B$ followed by ReLU followed by $C$.
The problem is that if we have $N$ categories, then $B$ needs to have $N^2$ rows! (Except if $q(X_1,X_2)$ is linear, in which case no
$ReLU$ is needed and things become trivial. For an explanation of this simple case, see end of Subsection \ref{first_transformer}).
Now, the third solution presented in Subsection \ref{third_transformer}, is an improvement of the first, where no matrix $B$ is needed, but
instead the linear out matrix is used in a clever way replacing the need for $B$.

Finally, the second solution uses self attention to obtain $q(X_j,X_i)$ for any pair $i,j\leq M$. This information is stored in the location part of 
the original vector space of the columns of $X$. The specific information about category pairs of adjacent tokens, that is $q(X_{i-1},X_i)$,
is then extracted with an ingenious trick with a skip connection followed by a ReLU with the appropriate bias. For that solution, there would be no need
for a fully connected layer, but we can still put in a matrix $B$ so as to have a compatible architecture for both approaches! We will have to take $N=M$
in order to have compatibility. Note that the first solution with a large matrix $B$, seems less
usefull then the third, which is similar in some sense.  In our mind, the race is between Second and Third solution, both can
be written :
$$\vec{Y}=C(ReLU(ATTENTION_*(Q,K,V)^t+X+\vec{bias})),$$
where $C$ is a one-row matrix.

One of the main problems, is that we defined our second solution without Softmax, but for the training to not get stuck into a gradient zero, we need Softmax or rescaling. Now, the Softmax will in general "ruin", the matrix \ref{attention2b}. Here is one idea, how to resolve this. The first step is to take $q^{\tt true}_{(.,.)}$ with values which are non-negative. Then we could take logarithm values of it, hence we would take $q=\ln(q^{\tt true}(.,.))$ and then the fact that Softmax takes exponnential function, would already be taken care
of this. The other problem, is that Sofmax will standartize every column of \ref{attention2b}. If the columns have enough elements, then, the sum of entries of column number $j$ would be approximately equal to the  expectation,
$E[q^{\tt true}(X,t)]$, where $t$ is not random and $X_j=t$.
So, for Softmax to not have negative effect, we need $E[q^{\tt true}(X,t)]$ for non-random $t$, to not depend on $t$.
This way, when we do the rescaling in each column of \ref{attention2b}, we get the same coefficient, which can be corrected later. With that correction,
even with Softmax, the Second Solution should be possible.

In the experiments though, we used rescaling implemented via a normalization layer, which prevents the gradients to become zero as well. 
		
Recall that $N$ is number of categories and $M$
is number of positions. Let us compare the two solutions in terms of the dimensions and formula for the different parts:
	
\bigskip

\begin{tabular}{c|c}
	\begin{tabular}{c}
		Attention represents \\
		category pairs \\
		$Q,K$ act only on categories
	\end{tabular}
	&\begin{tabular}{c}
		Attention represents \\
		neighboring token vectors \\
		$Q,K$ act only on positional encoding
	\end{tabular} \\\hline\hline
	&\\
	
	$Q^t=(q,0)\cdot X^t$
	&$Q^t=(0,q)\cdot X^t$\\
	
	$q\left[ N \times N \right]$
	&$q\left[ M \times M \right]$ \\
	
	$K^t=(I,0)\cdot X^t$
	&$K^t=(0,I)\cdot X^t$\\
	
	$k\left[ N \times N \right]$
	&$k\left[ M \times M \right]$ \\
	
	$v\left[(N+M) \times (N+M)\right]$
	&$v\left[(N+M) \times (N+M)\right]$\\
	
	No $B$ needed
	&$B\left[N^2 \times (N+M)\right]$ or no $B$ (third solution)
\end{tabular}
\bigskip

\subsection{Experiments}
We trained three systems with modified encoder layers, tailored towards the three proposed solutions. In these the correspending attention weights $k, q, v$ have access to either positional or input information. This was implemented by handling of the $X_C^T$ and $X_P^T$ separatelly. In addition we trained a system capable of learning all three solutions to investigate the natural choice of the transformer. In this system, the attention weights $k, q, v$ have access to both positional and input information. 

$X_C^T$ are one-hot encoded random samples from a categorical distribution $\sim Cat(N)$, where $N=10$ is the number of categories. $X_P^T$ are one-hot encoded M positions, where $M=50$ is the length of the sequence. The target $Y$ is generated by a function $q^{\tt true}(X_{i-1},X_i)$, where $q^{\tt true}(.,.)$ are random samples of a standard normal distribution. The models were trained in PyTorch with LBFGS optimizer using batch size of $1000$. The training was performed for $50$ iterations using Mean Squared Error (MSE) as a loss function. MSE from the last iteration was used to compare the performance of the models.

The attention mechanism was implemented as a single-head dot-product attention without Softmax. The output was then passed through a normalization layer followed by two linear layers with a ReLU activation in-between. The output of the second linear layer was the prediction of the model. The $LayerNorm$ normalization prevented the gradients from becoming zero.

The combined system capable of learning all three solutions was trained with the same setup as the individual models, but in four flavours: additional losses were introduced to lean the model towards each of the three proposed solutions as well as an unconstrained flavor using a single MSE loss.

\subsection{Results}
\begin{figure}[t]
\label{fig_log_mse}
\centering
\includegraphics[height=5cm]{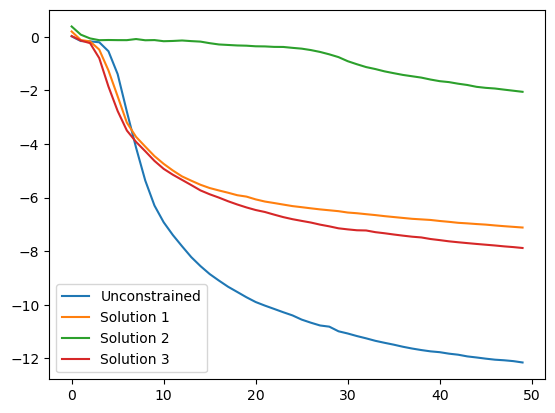}
\caption{Log-MSE for the combined model in four flavors}
\end{figure}

\begin{figure}[t]
\label{fig_combined}
\centering
\includegraphics[width=15cm]{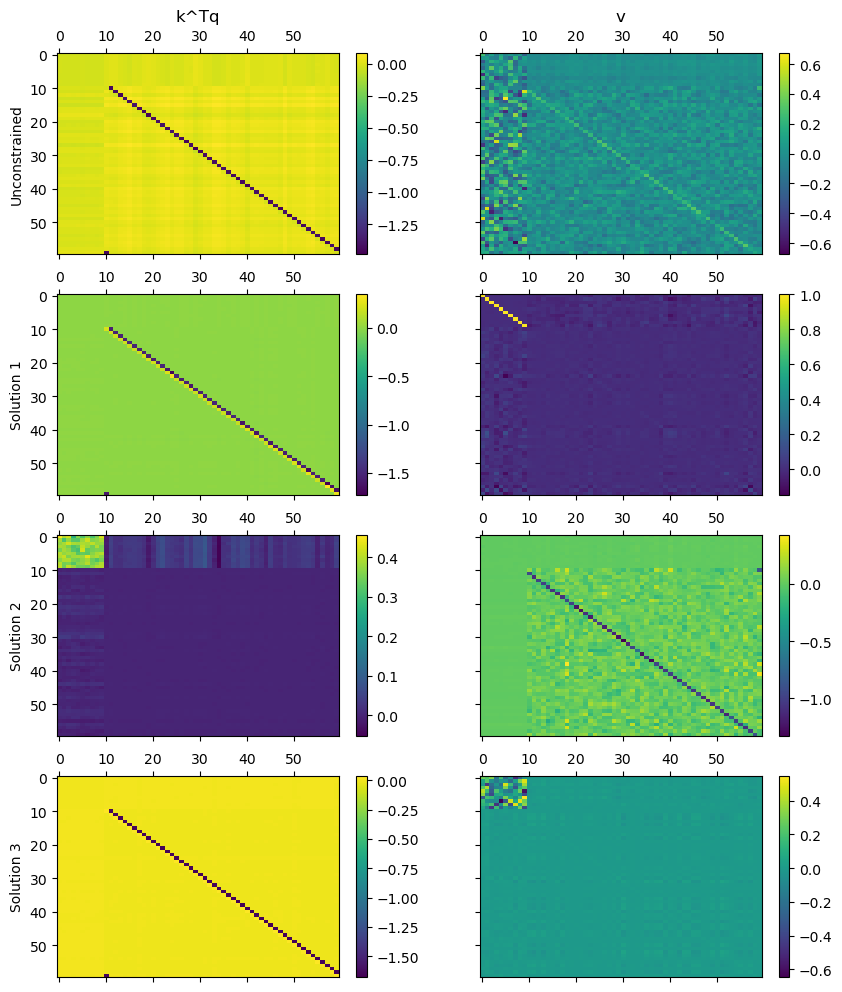}
\caption{Four flavours of the combined model}
\end{figure}

Results of the experiments are presented in Figures \ref{fig_log_mse} and \ref{fig_combined}. Figure \ref{fig_log_mse} shows $log-MSE$ of the running losses of the combined system in all variants. While the first and the third variants achieve much better performance in comparison to the second one. The unconstrained flavor performes the best by leveraging the combination of the third and the second solutions. Figure \ref{fig_combined} shows the attention weights of the combined system.

\begin{figure}[t]
\label{fig_sol_2}
\centering
\includegraphics[width=15cm]{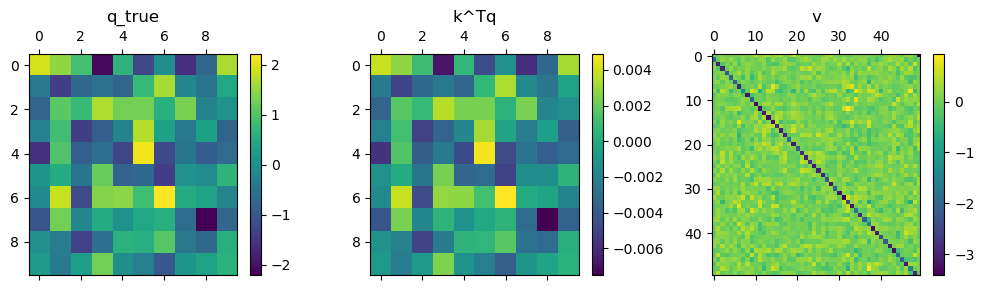}
\caption{Compare $q^{\tt true}$ with learned attention weights for the second solution}
\end{figure}

Figure \ref{fig_sol_2} shows the atentions weights of the system tailored exclusivelly towards the second solution. The $k, q$ are not able to capture the information about the positions $(X_{i-1},X_i)$, so, they store the $A_2$ matrix which is a scaled version of the $q^{\tt true}$ matrix.

\section{Conclusion}
Our experiments underscore the flexibility of the transformer architectures, i.e. its ability to perform tasks via multiple pathways and their combination. Key findings include the interesting symmetry in the attention mechanism allowing switching roles of $k^Tq$ and $v$.

This paper is an attempt to better understand the mechanics of the processing and transforming of sequential information within a single encoder layer. We studied a simple task of mapping the information from the adjacent positions into a real value. There are many more essential tasks in sequence processing, e.g. a memorization task. But, this might be the focus of the future work.

\bibliographystyle{unsrt} 
\bibliography{refs}

\end{document}